\pdfoutput=1 

 \def\isarxiv{1} 

\ifdefined\isarxiv
\documentclass[11pt]{article}

\usepackage[numbers]{natbib}

\else
\documentclass{article}
\usepackage{neurips_2022}
\fi

\usepackage[most]{tcolorbox}
\usepackage{hyperref}
\usepackage{tikz}
\usepackage{pgfplots}
\pgfplotsset{compat=newest}
\usepackage{microtype}
\usepackage{graphicx}
\usepackage{caption}
\usepackage{subcaption}
\usepackage{booktabs} 
\usepackage{authblk}

\usepackage{amsmath}
\usepackage{amsthm}
\usepackage{amssymb}
\usepackage{algorithm}
\usepackage{algpseudocode}
\usepackage{grffile}
\usepackage{wrapfig,epsfig}
\usepackage{url}
\usepackage{xcolor}
\usepackage{epstopdf}
\usepackage{bbm}

\allowdisplaybreaks

\usepackage{dsfont}
\usepackage[T1]{fontenc}
\usepackage{lmodern}
\usepackage{newtxtext}

\definecolor{applegreen}{HTML}{6fb600}

\newtcolorbox{promptbox}[2][]{%
  attach boxed title to top left
               = {yshift=-8pt},
  colback      = gray!10,          
  colframe     = gray!85,        
  colbacktitle = gray!85,          
  fonttitle    = \color{white}, 
  coltitle     = white,            
  title        = #2,              
  width=0.99\linewidth,              
  enhanced,                       
  breakable,
  before={\vspace{0pt}},
  after={\vspace{0pt}}
}

\newtcolorbox{promptbox2}[1][]{%
  colback      = gray!10,          
  colframe     = black,        
  width=0.99\linewidth,              
  enhanced,                       
  breakable,
  before={\vspace{0pt}},
  after={\vspace{0pt}}
}

\usepackage{silence}
\newcommand{\gopi}[1]{\begin{quote}{\textcolor{magenta}{\sf Gopi:} {\sl{#1}}} \end{quote}}
\newcommand{\jana}[1]{\begin{quote}{\textcolor{green}{\sf Jana:} {\sl{#1}}} \end{quote}}
\newcommand{\beibin}[1]{\begin{quote}{\textcolor{orange}{\sf Beibin:} {\sl{#1}}} \end{quote}}

\renewcommand{\gopi}[1]{}
\renewcommand{\jana}[1]{}
\renewcommand{\beibin}[1]{}

\newcommand{\DualAxisPlot}[4]{%
	\begin{tikzpicture}
		\begin{axis}[
			xlabel={\footnotesize Training Step},
			ylabel={\footnotesize \textcolor{blue!70}{Response Length}},
			axis y line*=left,
			axis x line=bottom,
			legend style={draw=none},
			width=\linewidth,
			height=0.75\linewidth,
			unbounded coords=discard,
            yticklabel style={rotate=45, 
            font=\small
            },
			xticklabel style={rotate=45, 
            font=\small,
            },
            scaled y ticks=false
			]
			\addplot[
			color=blue!70,
			mark options={scale=0.7},
			smooth,
			tension=0.1, 
			] table[
			x expr=\thisrow{_step} * #2,
			y=train/response_length,
			col sep=comma
			] {#1};
		\end{axis}
		
	\end{tikzpicture}%
}

\graphicspath{{./figs/}}

\ifdefined\isarxiv

\usepackage{tikz}
\usepackage{hyperref}  
\hypersetup{colorlinks=true,citecolor=blue,linkcolor=blue}
\usetikzlibrary{arrows}
\usepackage[margin=1in]{geometry}

\else

\usepackage{microtype}
\usepackage{hyperref}
\definecolor{mydarkblue}{rgb}{0,0.08,0.45}
\hypersetup{colorlinks=true, citecolor=mydarkblue,linkcolor=mydarkblue}

\fi

\makeatletter
\newcommand*{\RN}[1]{\expandafter\@slowromancap\romannumeral #1@}
\makeatother

\RequirePackage{lineno}
\usepackage{lineno}

\usepackage{tikz}
\usetikzlibrary{positioning} 
\usetikzlibrary{decorations.text}

\begin{document}

\ifdefined\isarxiv

\date{}

\title{On the Emergence of Thinking in LLMs I:\\ 
{\LARGE {\em Searching for the Right Intuition}}}
\author[1,*]{Guanghao Ye}
\author[2,*]{Khiem Duc Pham}
\author[3,*]{Xinzhi Zhang}
\author[4]{Sivakanth Gopi}
\author[4]{Baolin Peng}
\author[4]{Beibin Li}
\author[4]{Janardhan Kulkarni}
\author[5]{Huseyin A. Inan}
\affil[1]{MIT}
\affil[2]{Cornell University}
\affil[3]{University of Washington}
\affil[4]{Microsoft Research}
\affil[5]{Microsoft}

\renewcommand{\thefootnote}{\fnsymbol{footnote}}
\footnotetext[1]{Equal contribution.}
\renewcommand{\thefootnote}{\arabic{footnote}}

\else

\title{On the Emergence of Thinking in LLMs I: Searching for the Right Intuition}

\maketitle 
\fi

\ifdefined\isarxiv
  \maketitle
  \begin{abstract}
    Recent advancements in AI, such as OpenAI's new o models, Google's Gemini Thinking model, and Deepseek R1, are transforming LLMs into LRMs (Large Reasoning Models). Unlike LLMs, LRMs perform {\em thinking or reasoning} during inference, taking additional time and compute to produce higher-quality outputs. 
This work aims to discover the algorithmic framework behind  training LRMs. 
Approaches based on self-consistency, process reward modeling, AlphaZero, highlight that reasoning is a form of guided search. Building on this principle, we ask: what is the simplest and most scalable way to implement search in the context of LLMs?  

\smallskip

Towards answering these questions, we propose a post-training framework called Reinforcement Learning via Self-Play (RLSP). 
RLSP involves three steps: (1) supervised fine-tuning with human or synthetic demonstrations of the reasoning {\em process}, whenever possible (2) using an exploration reward signal to encourage diverse and efficient reasoning behaviors, and (3) RL training with an {\em outcome verifier} to ensure correctness while preventing reward hacking. Our key innovation is to decouple exploration and correctness signals during PPO training, carefully balancing them to improve performance and efficiency.

\smallskip

We perform empirical studies of the RLSP framework in the math domain, and show that
the models trained with the RLSP framework demonstrated improved reasoning abilities.
On Llama-3.1-8B-Instruct model the RLSP framework can boost performance by 23\% in MATH-500 test set;  
On AIME 2024 math problems, Qwen2.5-32B-Instruct improved by 10\% due to RLSP technique.

\smallskip

The more important finding of this work is that the models trained using RLSP technique, even with the simplest exploration reward that encourages the model to take more intermediate steps before arriving at a solution, showed several emergent behaviors such as backtracking, exploration of ideas, and verification. 
Furthermore, our framework enables such emergent behaviors across multiple model families, sizes, and domains.
These findings demonstrate that RLSP framework might be enough to enable the emergence of complex reasoning abilities in LLMs when scaled appropriately.

\smallskip

Lastly, we propose a theory as to why RLSP search strategy is more suitable for LLMs compared to previous approaches considered in the literature, inspired by a remarkable recent result that says that CoT {\em provably} increases computation power of LLMs, and hence reasoning, and these abilities grow as the number of steps in CoT \cite{li2024chain,merrill2023expresssive}. Our code is available at: \url{https://github.com/GuanghaoYe/Emergence-of-Thinking}.

  \end{abstract}


\else

\begin{abstract}

\end{abstract}

\fi

\clearpage

\begin{figure*}
    \centering
	\begin{tikzpicture}
    \node[text=black, fill=yellow!20, rounded corners, inner sep=2mm, text width=0.8\linewidth, align=center] (question) at (0,0) { 
\textbf{Question}: In rectangle $ABCD$, $AB=100$. Let $E$ be the midpoint of $\overline{AD}$. Given that line $AC$ and line $BE$ are perpendicular, find the greatest integer less than $AD$.
    };
    \node[ below=0.8em of question, ] (cot) {
			\begin{tcolorbox}[width=0.8\linewidth, title={ PPO with Outcome \& Length Reward}, colframe=black]
				\footnotesize

    \textbf{Answer}:

\textbf{\#\# Step 1:}
Draw a diagram of the given scenario, and we have a rectangle $ABCD$ with $AB=100$.

\textbf{\#\# Step 2:}
Let point $E$ be the midpoint of $\overline{AD}$. So, we can define $AE=ED=\frac{AD}{2}$.

...

\textbf{\#\# Step 27:}
Dividing both sides of the equation $10x^2=0$ by 10 gives $x^2=0$.

\textbf{\#\# Step 28:}
We know that $x^2=0$ is only true when $x=0$.

{\color{red}
\textbf{\#\# Step 29:}
However, since $ABCD$ is a rectangle, it has nonzero area, which means that $AD\ne0$.

\textbf{\#\# Step 30:}
Therefore, we must re-examine the steps to see where we went wrong.

\textbf{\#\# Step 31:}
Looking back at step 20, we can see that we derived two different expressions for $AC^2$, which is $2x^2$ and $-4x^2$.
}

...

\textbf{\#\# Step 57:}
The value of $100\sqrt{2}$ is approximately 141.42.

\textbf{\#\# Step 58:}
Since we are asked to find the greatest integer less than $AD$, we can use the value of $100\sqrt{2}$ we calculated in step 57.

\textbf{\#\# Step 59:}
The greatest integer less than $141.42$ is 141.

The final answer is: $\boxed{141}$.

    \end{tcolorbox}
    };
    \end{tikzpicture}
    \caption{The emergence of backtracking, self-correction in Llama-3.1-8B. During PPO training we incentivize the model to use more intermediate steps (length reward) before arriving at the solution. Full CoT trajectory in \autoref{sec:app-rlsp-emergence}.}
    \label{fig:llama_emergence_example}
\end{figure*}

\begin{figure*}
    \centering
	\begin{tikzpicture}
    \node[text=black, fill=yellow!20, rounded corners, inner sep=2mm, text width=0.8\linewidth, align=center] (question) at (0,0) {
    \textbf{Question}: The number \(\sqrt{104\sqrt{6}+468\sqrt{10}+144\sqrt{15}+2006}\) can be written as \(a\sqrt{2}+b\sqrt{3}+c\sqrt{5},\) where \(a, b,\) and \(c\) are positive integers. Find \(abc\).
    };
    \node[ below=0.8em of question, ] (cot) {
			\begin{tcolorbox}[width=0.8\linewidth, title={PPO with Outcome \& Length Reward}, colframe=black]
				\footnotesize

    \textbf{Answer}: To solve the problem, we start by assuming that the expression \(\sqrt{104\sqrt{6} + 468\sqrt{10} + 144\sqrt{15} + 2006}\) can be written in the form \(a\sqrt{2} + b\sqrt{3} + c\sqrt{5}\), where \(a, b,\) and \(c\) are positive integers. We need to find the values of \(a, b,\) and \(c\) such that:
    \[
    (a\sqrt{2} + b\sqrt{3} + c\sqrt{5})^2 = 104\sqrt{6} + 468\sqrt{10} + 144\sqrt{15} + 2006
    \]

...

    \[
    a = \sqrt{171} = 3\sqrt{19}
    \]

    {\color{red} Since $a$ must be an integer, we recheck our calculations and find a mistake.
    }
    
    Let's solve the system again:
    \[
    ab = 52,\quad 
    ac = 234,\quad
    bc = 72
    \]
    ...
    \end{tcolorbox}
    };
    \end{tikzpicture}
    \caption{The emergence of backtracking and self-correction in Qwen-2.5-32B-Instruct; full CoT reasoning trajectory in \autoref{sec:app-rlsp-emergence}.}
    \label{fig:qwen_emergence_example}
\end{figure*}

\begin{figure*}[htbp]
	\centering
	\begin{tikzpicture}

		\node[text=black, fill=yellow!20, rounded corners, inner sep=2mm, text width=0.8\linewidth, align=center] (question) at (0,0) { \textbf{Question:} The positive three-digit integer $N$ has a ones digit of $0$. What is the probability that $N$ is divisible by $4$? Express your answer as a common fraction.};

		\node[below=0.8em of question, anchor=north, xshift=-0.35\linewidth] (cot) {
			\begin{tcolorbox}[width=0.35\linewidth, title={ PPO with Outcome Reward Only}, colframe=black]
				\small
				\textbf{Step 1}:
				To determine the probability that the three-digit integer $N$ is divisible by $4$, we must first establish the total number of possibilities for a three-digit integer with a ones digit of $0$.
				
				\textbf{Step 2}:
				A three-digit number with a ones digit of $0$ is in the form of $\overline{ab0}$, where $a$ and $b$ are digits that can take on values from $1$ to $9$, since the hundreds digit cannot be $0$. \textcolor{red!70}{Therefore, there are $9$ options for $a$ and $9$ options for $b$, giving a total of $9 \cdot 9 = 81$ possible numbers.}
				
				\textbf{Step 3}:
				For a number to be divisible by $4$, the last two digits must be divisible by $4$. Since the last digit is $0$, we only need to consider the possible values of $b$ that make the number $\overline{b0}$ divisible by $4$.
				
				\textbf{Step 4}:
				We can systematically check values of $b$ from $0$ to $9$ to determine which values make $\overline{b0}$ divisible by $4$. By doing so, we find that the values of $b$ that make $\overline{b0}$ divisible by $4$ are $4$ and $8$.
				
				\textbf{Step 5}:
				Thus, there are $2$ possible values for $b$ that make the number $\overline{b0}$ divisible by $4$, and each of these has $9$ options for $a$. Therefore, there are $9 \cdot 2 = 18$ possibilities that are divisible by $4$.
				
				\textbf{Step 6}:
				To find the probability that $N$ is divisible by $4$, we divide the number of possibilities that are divisible by $4$ by the total number of possibilities, which is $\frac{18}{81} = \frac{2}{9}$.
				
				The final answer is: $\boxed{\frac{2}{9}}$
			\end{tcolorbox}
		};
		\node[below=0.8em of question, anchor=north, xshift=0.15\linewidth] (rlsp) {
			\begin{tcolorbox}[width=0.6\linewidth, title={ PPO with Outcome \& Length Reward}, colframe=black]
				\small
				\textbf{Step 1}: Understand the conditions for divisibility by 4
				
				A number is divisible by 4 if the number formed by its last two digits is divisible by 4.
				
				\textbf{Step 2}: Recognize the possible last two digits for N
				
				Since N is a three-digit number and its ones digit is 0, we only need to consider the last two digits which can range from 00 to 90, in increments of 10.
				
				\textbf{Step 3}: List the possible last two digits of N
				
				The possible last two digits are 00, 10, 20, 30, 40, 50, 60, 70, 80, and 90.
				
				\textbf{Step 4}: Determine how many of these are divisible by 4
				
				Among 00, 10, 20, 30, 40, 50, 60, 70, 80, and 90, the numbers that are divisible by 4 are 00, 20, 40, 60, and 80.
				
				\textbf{Step 5}: Count the total number of possibilities
				There are 10 possibilities in total.
				
				\textbf{Step 6}: Calculate the probability that N is divisible by 4
                
				The probability that N is divisible by 4 is the number of successful outcomes (numbers divisible by 4) divided by the total number of outcomes.
				
				\textbf{Step 7}: Calculate the probability
                
				There are 5 successful outcomes (00, 20, 40, 60, 80) and 10 total outcomes. \textbf{So the probability is 5/10 which simplifies to 1/2}.
				
				\textbf{Step 8}: Consider the divisibility rule for 4
				
				\textit{... omitted ...}

				\textbf{Step 16}: Realize that the probability that N is divisible by 4 is still 1/2
				
				\textbf{This is because 1/2 of the numbers from 00 to 90 are divisible by 4, regardless of the other digits in the number.}
				
				\textbf{Step 17}: Consider the fact that the hundreds digit does not affect the probability
				
				\textit{... omitted ...}

				\textbf{Step 22}: \textcolor{purple}{Understand that this confirms our previous conclusion. 
                				We have already determined that 5 of the numbers from 00 to 90 are divisible by 4.}
				
				\textbf{Step 23}: Realize that the probability that N is divisible by 4 is 1/2
                
				Since 5 of the 10 numbers from 00 to 90 are divisible by 4, \textbf{the probability is 5/10, which simplifies to 1/2.}
				
				The final answer is: $\boxed{\frac{1}{2}}$
			\end{tcolorbox}
		};
		
	\end{tikzpicture}
	\caption{A side-by-side comparison of regular CoT and CoT after RLSP training. Both experiments use Llama-3.1-8B-Instruct model without any additional SFT. Observe that the model with length reward, which incentivizes more intermediate steps, exhibits emergent behaviors such as verification (step 22) and consideration of alternative possibilities (texts in bold). The full reasoning trajectory in \autoref{sec:app-rlsp-emergence} shows that the model tries 3 different approaches to solve the problem and recognizes that they all lead to the same answer.}
    \label{fig:emergence_example}
\end{figure*}

\clearpage

\section{Introduction}

With the release of o-series of models from OpenAI \cite{openai-o1-mini,openai-o1}, Gemini Thinking model from Google \cite{gemini-2}, and Deepseek R1 \cite{deepseek-r1}, LLMs are rapidly evolving into thinking machines,  now referred to as LRMs (Large Reasoning Models).
The key technical difference between LLMs and LRMs is the ability of LRMs to do {\em thinking} during the inference, which we define as the ability to take more time and compute during (inference) with the goal of producing a higher quality output to a given input, 
which is a fair definition of thinking process in all systems capable of reasoning both artificial and biological.

The main goal of this work is to discover the computational process behind reasoning or thinking in the context of LLMs.
There have been several attempts in the past towards understanding this process, ranging from techniques such as self-consistency \cite{wang2022self, liu2023pre}, (automated) process reward modeling (PRM) \cite{uesato2022solving,lightman2023let,snell2024scaling,beeching2024scalingtesttimecompute}, and adapting AlphaZero style search for LLMs \cite{feng2023alphazero,chen2024alphamath,trinh2024alphageometry,alphaproof}. 
All of these techniques have one principle in common: reasoning or thinking is {\em some form of search}.
We build on the principle of defining thinking as a form of guided search and ask the question:

\smallskip
{\em What is the simplest and most scalable framework for training LLMs that leads to the emergence of thinking or search behavior?}

\subsection{Our Contributions}

\paragraph{The RLSP Framework}
Towards answering these questions, we propose a {\em post-training} framework called Reinforcement Learning via Self-Play (RLSP)\footnote{RLSP name is inspired by the quote "Play is the highest form of research'' attributed to Albert Einstein and the self-play technique used in AlphaZero systems}.
Our framework is a natural extension and generalization of the RLHF \cite{ouyang2022training} and RL from Constitutional AI feedback \cite{bai2022constitutional} frameworks and consists of three simple steps: 

\begin{itemize}

\item \textbf{Step 1:} If high-quality demonstrations of the {\em process of thinking} are available, either via human annotations or synthetic traces constructed via tree search procedures, do SFT (supervised fine-tuning) on the demonstration dataset.

\item \textbf{Step 2:} Use an {\em exploration reward} that is {\em independent} of the correctness of the solution to
implicitly encourage {\em diverse} search behaviors such as  backtracking,   consideration of alternative possibilities, verification, etc. 

\item \textbf{Step 3:} Do RL (reinforcement learning) with PPO (proximal policy optimization, \cite{schulman2017proximal}) as the training algorithm, using an {\em outcome verifier} that gives an unambiguous {\em binary signal of the correctness} of the solution.
\end{itemize}

In this work, we focus on domains where outcome verification is possible (and easier) during training.
Our key insight is that any reward engineering in RL training should encourage the model to synthetically create {\em novel} CoT data that it can learn from during the PPO training; that is, it incentivizes self-play over new CoT reasoning traces.
A simple way to implement this is to decouple the exploration reward signal that encourages search behavior from the correctness of the solution, and incentivize it.
During PPO training, we do a careful weighing of the score from the outcome verifier and the score from exploration reward model for the optimal performance.
By design, both  components of the reward signal mitigate issues such as reward hacking and overfitting to the training data.
More importantly, both signals provide minimal but essential feedback for the model to learn better with scale consuming more compute and data.

In \autoref{sec:theory} we propose an argument supporting these intuitions and how we arrived at the RLSP framework guided by some remarkable recent results \cite{li2024chain,merrill2023expresssive}.

\paragraph{Empirical Performance Analysis: {\em RLSP search leads to better reasoning in LLMs.}} 

\begin{figure}[htbp]
\begin{center}
 \centerline{\includegraphics[width=0.7\columnwidth]{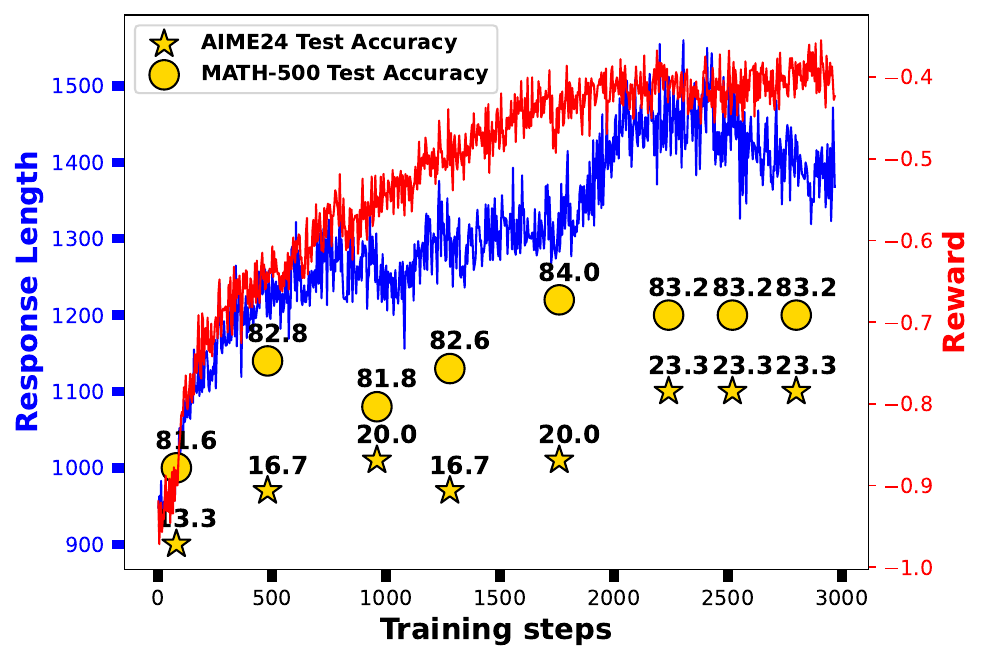}}
\caption{Reward, response length and AIME24 accuracy during RL training with the PPO algorithm using the simplest exploration reward: reward thinking more. 
The increase in response length is not sufficient but necessary for search behavior and better reasoning, even in a strict theoretical sense \cite{merrill2023expresssive}. The base model is Qwen2.5-32B-Instruct; no SFT or special prompting were employed during training and inference.}
\label{fig:qwen_32b_aime}
\end{center}
\end{figure}

We empirically evaluate the improvement in reasoning abilities of models trained using the RLSP framework in the math domain.

On Llama models, the RLSP framework can boost performance by 23\% on the MATH dataset \cite{hendrycks2021measuring}; On AIME 2024 math problems, Qwen2.5-32B-Instruct improved by 10\% due to RLSP technique as shown in Figure~\ref{fig:qwen_32b_aime}. 

Moreover, under the {\em same token budget}, the self-consistency (SC) accuracy of
Llama-3.1-8B-Instruct on MATH-500 is 61.8\%, whereas the same model trained via RLSP achieves 73.4\%.
Our empirical findings show that enabling guided search behavior leads to significant gains in the reasoning abilities of LLMs compared to CoT and SC.

\paragraph{Emergent Properties and Behavioral Analysis}
The most interesting contribution of our work is understanding the emergent properties of models trained via RLSP, and the various ablation studies we perform to delineate the role played by SFT, RL with or without exploration reward, and the impact of base models and the pretraining data.

We show that even with no SFT but equipped with minimal exploration reward, such as incentivizing longer generations that explicitly output all intermediate steps of a reasoning trajectory, {\em all} models in {\em both} coding and math domains learn several interesting search behaviors. 
They show various emergent properties such as exploring alternative possibilities, cross-verification, backtracking, and self-correction, which are some of the well-known meta search strategies.
See \autoref{fig:llama_emergence_example}, \autoref{fig:qwen_emergence_example}, and \autoref{fig:emergence_example}; \autoref{sec:app-rlsp-emergence} contains several more examples of emergent behaviors of models  just using the generation length reward.

On the other hand, if there is no exploration reward but pure PPO training based on binary reward from the verifier, search behavior {\em only} emerged in Qwen2.5-7B-Instruct model on math domain (and not in coding domain for the same base model).

These results indicate, not surprisingly, that emergent properties of the models trained via pure RL can vary widely based on the pretraining data and the choice of base models. 
In contrast, {\em RLSP enables emergent  search behaviors across multiple model families, sizes, and
domains.}
Thus, to train frontier models at scale, RLSP offers a smoother and more efficient framework to equip LLMs with sophisticated search behaviors that can lead to improved reasoning.

\paragraph{Remark 1} A word about our choice of words. We use the phrase "thinking behavior'' to mean the ability of LLMs to take more time and compute with the goal to produce a higher quality output to a given input, use it interchangeably to mean search behavior or reasoning. This is based on our hypothesis that search enables thinking which in turn should lead to better reasoning. However, technically speaking, search/thinking behavior is necessary for reasoning \cite{li2024chain,merrill2023expresssive} but not sufficient. 

\paragraph{Remark 2} We call a behavior emergent if the model does not exhibit that behavior in regular CoT without any special prompting techniques. 
However, it should be noted that the behaviors we call emergent in this paper such as backtracking or self-correction are implicitly present in pretraining data, although it is less frequent than standard step-by-step reasoning. 
See \autoref{sec:theory} and the discussion in \autoref{sec:method_ppo} for some mathematical justification of why standard CoT or pure RL may not always show behaviors such as backtracking or self-correction.
Finally, we also notice that emergent behaviors, although in principle should lead to better reasoning, may not always result in correct solutions; see \autoref{sec:app-rlsp-emergence} for an example.
We speculate that this could be due to small model sizes, data, and compute we use in our experiments.

\paragraph{Remark 3} We are aware of several other works that are published in the last 2-3 weeks, including the Deepseek r1 report, which have  findings that are related to our work.
This is a concurrent work (that is already under ICML submission) and hence we do not compare to those very recent papers.
However, we believe that our work is more principled and has many new findings.

\subsection{A Theory of RLSP} 
\label{sec:theory}
We propose a theory to explain how we arrived at the RLSP search strategy and how it can lead to continuous self-improvement. 

The guiding principle behind RLSP is that any RL training technique should incentivize the model to synthetically create {\em novel} CoT reasoning trajectories that are not already present in the training data, and learn from it.
Our intuition comes from a recent elegant mathematical result that states CoT can {\em provably} enhance the computational power of transformers \cite{li2024chain,merrill2023expresssive}.
Broadly speaking, \cite{merrill2023expresssive, li2024chain} argument says that the {\em length} of  chain-of-thought trace impacts its reasoning power, and more intermediate steps lead to more computational power under standard computational complexity assumptions.
Recall that CoT is an empirical implementation of 
$\arg \max P_\theta(\text{answer, rationale}~|~\text{problem})$ 
for an autoregressive LLM parameterized by $\theta$.

Suppose we assume that as the difficulty of a reasoning problem increases, arriving at the {\em right rationale or intuition} necessary to solve the problem becomes harder.
In particular, commonly occurring ideas in the pretraining data, which is what standard CoT finds, fail to lead to the correct solution.
Then it is natural to train the model to {\em search} over the space of rationales thus maximizing $P_\theta(\text{correct rationale, trajectory over rationales}~|~\text{problem})$, which can be interpreted as CoT in the space of trajectories over rationales.
However, such trajectories may not be present in training data.
Here, it is beneficial to think of settings where the problems are so difficult that no human can solve it. 
Taking cue from \cite{merrill2023expresssive, li2024chain},  we can still design reward signals that encourage the model to use more intermediate steps as the problem difficulty increases and explore diverse rationales to solve the problem.
This was our motivation to design exploration reward signal in RLSP.
During the RL training most of the reasoning trajectories fail to lead to the correct answer; yet the model gets a small reward for the exploration. 
However, when it eventually finds the correct answer via a long reasoning trajectory, it gets the full reward.
Thus, exploration and response length reward signals in RLSP encourage the model to generate all intermediate steps (CoT) to arrive at a solution, thus synthetically creating the CoT trajectories to learn from during PPO.

In nutshell, {\em RLSP enables models to generate new CoT data via self-play.}
Since we already know that CoT improves reasoning abilities of LLMs, RLSP in principle can keep improving the reasoning abilities as long as there are sufficiently diverse new problems to solve!

These discussions should also give an intuition to the reader why an SFT dataset over the demonstrations of reasoning traces can only help the model during RL training, and in fact can be viewed as created by some search process either by humans or other techniques such as tree search.
Thus, SFT over reasoning traces and RL should be considered as working together to instill search behavior in LLMs.
However, in the limit, RL training with exploration and outcome reward signals enables the models to continuously self-improve by creating synthetic CoT traces.

Finally, we note that many of the emergent behaviors of models trained using RLSP give some validation to this theory. For example, in \autoref{fig:emergence_example} the model searchers over multiple rationales (and within each rationale it does CoT) before verifying that all of them lead to the same answer.
Similar behaviors are also seen in other examples given in \autoref{sec:app-rlsp-emergence}.

\section{Details of the RLSP Framework}

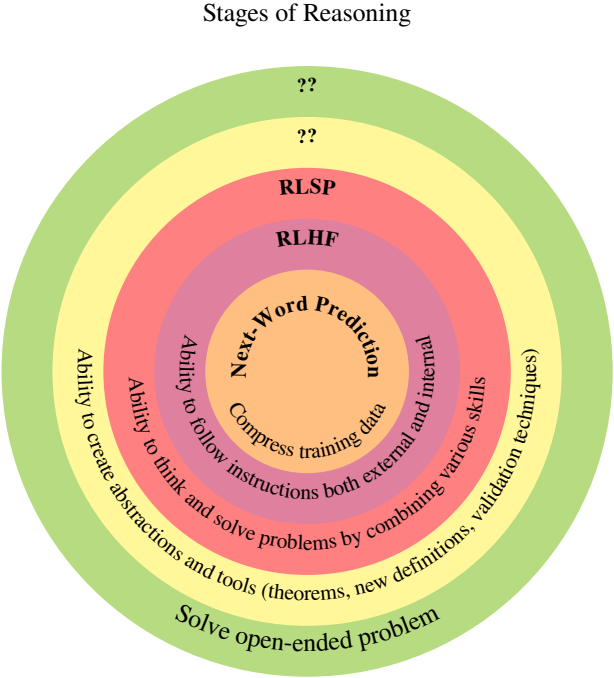
\begin{figure}[htbp]
\centering
\resizebox{0.5\columnwidth}{!}{%
    \begin{tikzpicture}[node distance=0.5cm and 1cm, font=\normalsize]
    \fill[applegreen!50] (0,0) circle (6cm); 
    \fill[yellow!50] (0,0) circle (5cm); 
    \fill[red!50] (0,0) circle (4cm); 
    \fill[purple!50] (0,0) circle (3cm); 
    \fill[orange!50] (0,0) circle (2cm);  

    \draw[decorate, 
    decoration={
        text effects along path, 
        text={Next-Word Prediction},
        text align=center,
        text effects/.cd,
        text along path,
        characters={font=\large\bfseries} 
    }] (0,-1.2) arc (270:-90:1.2cm);
    
    \draw[decorate, 
    decoration={
        text effects along path,
        text={Compress training data},
        text align=center,
        text effects/.cd,
        text along path,
        characters={font=\large} 
    }] (0,1.7) arc (90:450:1.7cm);

    \draw[decorate, 
    decoration={
        text effects along path, 
        text={RLHF},
        text align=center,
        text effects/.cd,
        text along path,
        characters={font=\large\bfseries} 
    }] (0,-2.5) arc (270:-90:2.5cm); 

    \draw[decorate, 
    decoration={
        text effects along path,
        text={Ability to follow instructions both external and internal},
        text align=center,
        text effects/.cd,
        text along path,
        characters={font=\large} 
    }] (0,2.5) arc (90:450:2.5cm); 

    \draw[decorate, 
    decoration={
        text effects along path, 
        text={RLSP},
        text align=center,
        text effects/.cd,
        text along path,
        characters={font=\large\bfseries} 
    }] (0,-3.5) arc (270:-90:3.5cm);
    \draw[decorate, 
    decoration={
        text effects along path,
        text={Ability to think and solve problems by combining various skills},
        text align=center,
        text effects/.cd,
        text along path,
        characters={font=\large} 
    }] (0,3.5) arc (90:450:3.5cm); 

    \draw[decorate, 
    decoration={
        text effects along path, 
        text={??},
        text align=center,
        text effects/.cd,
        text along path,
        characters={font=\bfseries}
    }] (0,-4.5) arc (270:-90:4.5cm); 
    \draw[decorate, 
    decoration={
        text effects along path,
        text={Ability to create abstractions and tools (theorems, new definitions, validation techniques)},
        text align=center,
        text effects/.cd,
        text along path,
        characters={font=\large} 
    }] (0,4.5) arc (90:450:4.5cm);

    \draw[decorate, 
    decoration={
        text effects along path, 
        text={??},
        text align=center,
        text effects/.cd,
        text along path,
        characters={font=\bfseries}
    }] (0,-5.5) arc (270:-90:5.5cm); 
    \draw[decorate, 
    decoration={
        text effects along path,
        text={Solve open-ended problem},
        text align=center,
        text effects/.cd,
        text along path,
        characters={font=\Large} 
    }] (0,5.5) arc (90:450:5.5cm);

    \node at (0, 7) {\Large Stages of Reasoning};
\end{tikzpicture}
}
\caption{A possible training paradigm for the emergence of complex reasoning process. In this work, we propose RLSP framework to enable the thinking process.}
\label{fig:stagesofreasoing}
\end{figure}

In this section, we describe full RLSP framework that we used to train our best reasoning models.
In \autoref{sec:can_pure_rl_lead_to_thinking}, we do ablation studies to understand how individual steps of RLSP change the thinking behavior of models, and show why all 3 steps may be the smoothest way to empower search behavior in LLMs. However, SFT step may be optional with proper reward engineering during RL stage.

RLSP is a post-training framework for LLMs in similar spirit as RLHF. While RLHF encourages instruction following and teaches human preferences, RLSP encourages reasoning in model responses.
We give a meta hierarchy for model training in \autoref{fig:stagesofreasoing} which shows how each of these frameworks encourage different abilities in language models and what emergent abilities can future frameworks achieve.
Our framework consists of following steps.

\subsection{SFT of the Base Model}\label{sec:method_sft}
In the first step of RLSP framework, we perform a supervised fine-tuning using cross-entropy loss.
A high-quality SFT dataset should contain demonstrations of the {\em thinking process} that incorporates typical reasoning principles such as backtracking, abandoning a reasoning chain, self-verification, etc.
We note that SFT dataset need not be exhaustive in terms of all possible reasoning traces; 
In simpler terms, this step can be thought of as studying chapters in a textbook or attending lectures on a particular topic before asking the student to solve homework problems.
In fact, this step is not even necessary, but helps in  (RL) training models that exhibit better search behavior as we will see in coming sections.

SFT datasets can be constructed through various methods: 1) Human demonstrations. 2) Depth First Search (DFS) traversal of MCTS or other tree search techniques 3) Synthetic data via agentic workflows or using a thinking model that is already trained. 
Although the quality of data produced by each of these steps can vary, with proper curation one could hope to yield high-quality demonstration data.
Extensive research on the relative efficacy of these techniques is beyond the scope of this paper, and we leave it as an open problem.
In our experiments, we create an SFT dataset using a filtered version of publicly available reasoning traces QwQ-LongCoT \cite{qwq-longcot}.

\subsection{Reward Function}\label{sec:method_creativity_reward}

The most important component of RL training is establishing the reward function $\mathcal{R}$. Suppose for a given prompt $q$ the model outputs $o$. Suppose we have an outcome verifier $\mathrm{Ver}$ which objectively decides whether the model response $o$ given prompt $q$ is correct or incorrect (which can be done in math domain if we know the final answer for a problem and in coding domain by running the code on a few test cases). We will also utilize an \emph{exploration reward} $\mathcal{R}_{\mathrm{ex}}(q,o)$ which judges the effort and creativity shown by the response $o$ in answering the prompt $q$.
During training, the output reward signal $\mathcal{R}(q,o)$ consists of two components:
\begin{align}
    \mathcal{R}(q,o) = \alpha \cdot \mathds{1}\left[\mathrm{Ver}(q,o) =\mathrm{True}\right] + (1-\alpha) \cdot \mathcal{R}_{\mathrm{ex}}(q,o).
\label{eq:reward}
\end{align}
The key insight in our work compared to most previous approaches based on PRM is to give a reward signal independent of the output correctness that encourages desirable properties of the process of reasoning.
This is related to concept of auxiliary rewards in RL, we refer the reader to \cite{jaderberg2016reinforcement} and references therein.
Unlike PRM, the exploration reward does not {\em directly} measure the progress the policy model is making towards the outcome, but rather meta-properties of the reasoning trajectory that increases the success probability, as discussed in \autoref{sec:theory}.
Thus, we reward the process itself instead of process reward modeling (PRM). 
Moreover, in the beginning of RL training, when the reward signal from the outcome verifier is very sparse, the exploration reward serves as a dense signal guiding the model towards longer and better reasoning trajectories.

To prevent reward hacking of the exploration reward signal, we carefully balance the correctness signal from the outcome verifier and exploration reward.
In practice $\alpha$ is a hyperparameter that needs to be tuned  and decayed over time,  but we set it as 0.8 in our experiments for simplicity; 
that is bulk of the reward signal comes purely from the outcome verifier which makes the learning process more challenging but enables better generalization.

In our experiments, we do not train an exploration reward model; instead, we experiment with two simple ways to assign this exploration reward:

\begin{itemize}

\item {\em Based on length of response.} That is, $\mathcal{R}_{\mathrm{ex}}(q,o)=-\frac{C}{|o|}$ where $|o|$ is the number of tokens in the response and $C>0$ is some constant. So longer responses, which explicitly output all the intermediate steps taken to arrive at a solution, get a higher reward. This is based on the intuition that for difficult problems the reasoning effort is correlated with length of the model response, which is made precise by the work of \cite{merrill2023expresssive}.
During the discussion of PPO training (see Eq. \eqref{eq:whyexplore}), we will see another motivation for adding this reward explicitly.
Just as importantly, this is a simple and objective reward to implement.

As we saw earlier, even with this simple exploration reward, models show several emergent behaviors after PPO training \autoref{fig:emergence_example}, \autoref{fig:qwen_emergence_example}, \autoref{fig:llama_emergence_example}.
However, in many cases, this exploration reward signal is not enough as the models learn to hack the reward signal by repeating responses, necessitating a more nuanced reward signal.

\item {\em By using LLM-as-a-judge} to score $(q,o)$ based on the creativity, reasoning effort and other merits as judged by an independent LLM (while ignoring the correctness of the response). In our paper, we use GPT-4o model as the judge, see the prompt we use in \autoref{sec:prompts}.
We note that we use GPT-4o model as a judge as our base models are rather small; however, if the base model is a frontier model then the policy  and the judge models need not be different.

\end{itemize}

One could use constitutional AI approach to train an exploration reward model using RL \cite{bai2022constitutional}, which can be more easily adapted to new reasoning domains. 
Here, one can formulate a constitution of what constitutes novel or creative reasoning process for that specific domain, and steer the model towards those desired behaviors.
We leave this as an interesting future research direction.

\subsection{RL Training with PPO}\label{sec:method_ppo}
We use the PPO algorithm \cite{schulman2017proximal} to train our policy model.
Our implementation of the PPO algorithm for training the policy is similar to the setup used in RLHF fine-tuning using PPO \cite{ouyang2022training}. Suppose the model response can be decomposed into tokens as $o=o_1,o_2,\dots,o_T$ with $o_T=\mathrm{EOS}$ being the end-of-string token.
The loss function in PPO is given by:
\[
L^{\text{PPO}}(\theta) = \mathbb{E}_t \left[ \min\left(\rho_t(\theta) \hat{A}_t, \text{clip}(\rho_t(\theta), 1 - \epsilon, 1 + \epsilon) \hat{A}_t\right) \right]
\]
where $\rho_t(\theta)=\frac{\pi_\theta(o_t|q,o_{<t})}{\pi_{\theta_{\mathrm{old}}}(o_t|q,o_{<t})}$ and $\hat{A}_t$ is the GAE estimate for the advantage function given by:
$\hat{A}_t=\delta_t+(\gamma \lambda)\delta_{t+1}+\dots+(\gamma\lambda)^{T-t} \delta_{T}$ where $\delta_t=r_{t} +\gamma V_\phi(q,o_{<(t+1)}) - V_\phi(q,o_{<t})$ and $r_t$ is the per-token reward at step $t$ and $V_\phi(q,o_{<t})$ is the value function predicted by critic model which is trained along with the policy model using squared loss with target given by reward-to-go $\hat{R}_t=r_t+\gamma r_{t+1}+\dots+\gamma^{T-t}r_T$. We use the implementation of PPO from OpenRLHF framework \cite{hu2024openrlhf} for our experiments and set $\gamma=1$ and $\lambda=0.95$. 
As in previous works, we also use a KL penalty term to stop the model from drifting too far from the base SFT model ($\pi_{\theta_{\mathrm{SFT}}}$). 
Therefore the per-token reward is given by:
$$r_t = \mathcal{R}(q, o)\cdot \mathds{1}(o_t=\mathrm{EOS}) - \beta \log\left(\frac{\pi_{\theta_{\mathrm{old}}}(o_t|q,o_{<t})}{\pi_{\theta_{\mathrm{SFT}}}(o_t|q,o_{<t})}\right).$$
Note that the expected total reward is 
\begin{align} 
&\mathbb{E}_{o \sim \pi_{\theta_{\text{old}}}(\cdot|q)}\left[\sum_{t=1}^T r_t\right] =  \mathbb{E}_o[\mathcal{R}(q,o)] 
 - \beta \sum_{t=1}^T D_{\mathrm{KL}}\left(\pi_{\theta_{\text{old}}}(\cdot|q,o_{<t}) \|\ \pi_{\theta_{\text{SFT}}}(\cdot|q,o_{<t})\right). \label{eq:whyexplore}
\end{align}
Since the KL divergence at each step is non-negative, the KL penalty term leads to lower reward for longer responses which could prevent the model from learning to produce long reasoning responses. {\em The exploration reward compensates this by rewarding longer responses.} In our experiments, adding the simple length based exploration reward consistently leads to longer responses with detailed CoT and better reasoning behavior in the model. But without the exploration reward, it works for some settings and doesn't work for some others.

\section{Can Pure RL Lead to Thinking Behavior?}
\label{sec:can_pure_rl_lead_to_thinking}
An immediate question that arises from the RLSP framework is: Are SFT and exploration reward signals necessary or can LLMs learn to search directly with pure PPO training on the binary output of a verifier?
Recall that \autoref{eq:whyexplore} sheds some light on this question in a theoretical sense.
To empirically answer these questions, we perform following set of experiments on Llama and Qwen models.
In these experiments we use response length as an objective metric to measure the search behavior.
We will not concern ourselves with accuracy in this section. 

\begin{figure}[h!]
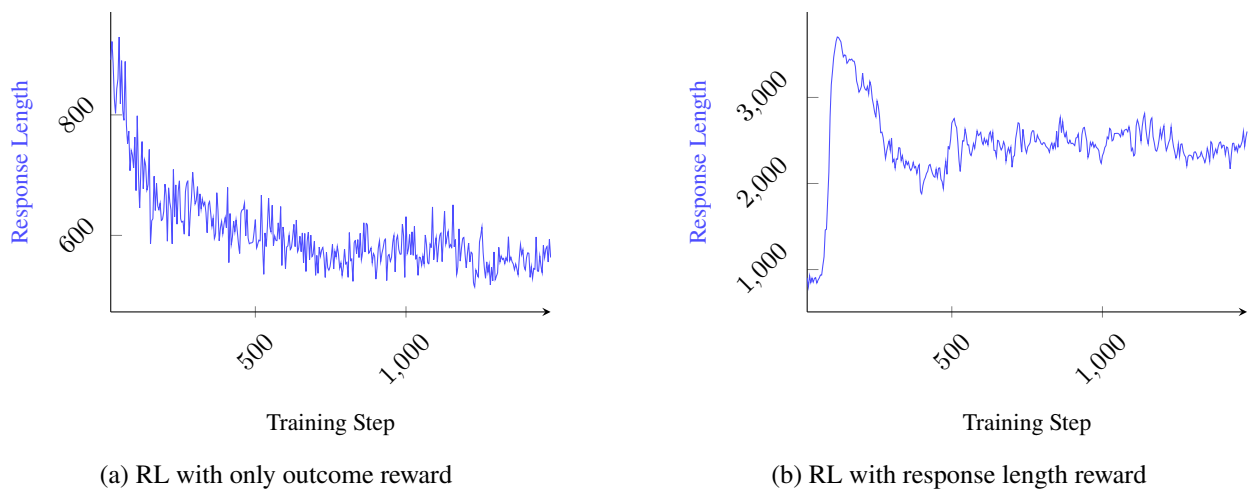

    \centering
    \begin{subfigure}[b]{0.45\textwidth}
        \centering
        \DualAxisPlot{figs/data/llama_8b_instruct_no_length_penalty.csv}{4}{200}{1500}
        \caption{RL with only outcome reward}
        \label{fig:llama_wo_length_p}
    \end{subfigure}
    \hfill
    \begin{subfigure}[b]{0.45\textwidth}
        \centering
        \DualAxisPlot{figs/data/llama_8b_instruct_length_penalty.csv}{4}{50}{450}
        \caption{RL with response length reward}
        \label{fig:llama_w_length_p}
    \end{subfigure}

 \caption{Response length during RL training of Llama-3.1-8B-Instruct on MATH dataset.}
 \label{fig:llama_response_len}
\end{figure}

\paragraph{Llama  Models}
Starting from Llama-3.1-8B-Instruct \cite{llama-31-8b} as the base model on which we perform no SFT, we do a) PPO training on the training dataset of MATH with no exploration reward of any kind; that is $\alpha = 1$ in Eq. \eqref{eq:reward}.
b) PPO training on the training dataset of MATH with the creativity reward proportional to length: $R_{cr}(q,o) \propto-\frac{1}{|o|}$.
Our findings are in \autoref{fig:llama_response_len}.

Our results indicate that on Llama-3.1-8B-Instruct models pure RL with only outcome reward does not lead to search behavior. 
It is also reflected in the solutions produced by the model, which show no emergent search behavior, see \autoref{fig:emergence_example}.
On the other hand, even when we give a weak signal of exploration reward that is proportional to the response length, the model exhibits several interesting search behaviors; see \autoref{fig:llama_emergence_example},  \autoref{fig:qwen_emergence_example}, \autoref{fig:emergence_example} and \autoref{sec:app-rlsp-emergence}.
{\em We want to emphasize that the increase in response length in this case is expected but what is surprising is the emergence of new search behaviors such as backtracking and verification, which models do not seem to perform in standard CoT.}

\begin{figure}[h!]
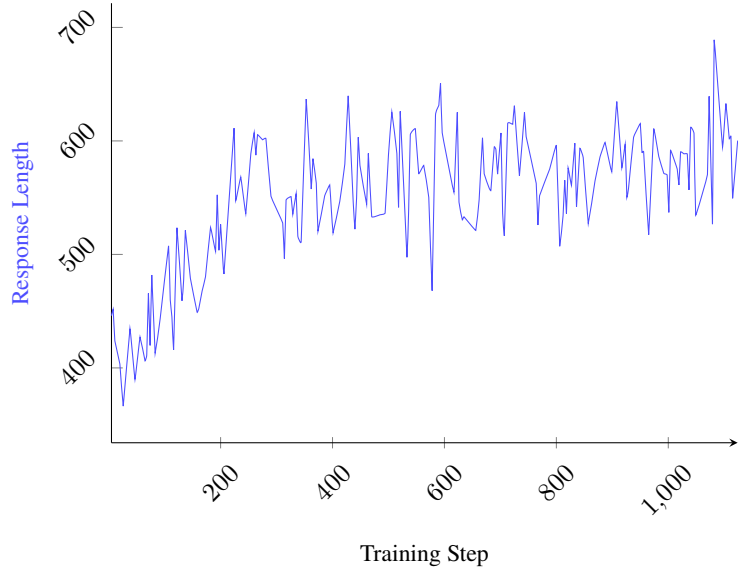

    \centering
    \begin{subfigure}[b]{0.6\textwidth}
        \centering
        \DualAxisPlot{figs/data/rl-math-qwen-2.5-7b_export.csv}{1}{50}{450}
        \label{fig:qwen_rl_math}
    \end{subfigure}

 \caption{Response length during RL training of Qwen2.5-7B-Instruct model with only outcome reward in math domain.}
 \label{fig:qwen_result_math}
\end{figure}

\paragraph{Qwen Models}
We also carried out a similar experiment with Qwen2.5-7B-Instruct model in math domain.
The result is presented in \autoref{fig:qwen_result_math}.

Interestingly, we observe that Qwen2.5-7B-Instruct model does demonstrate an increase in the response length with pure RL training using only outcome reward.
Thus the base model and the pretraining data used for training them can have significant impact on the emergent behaviors.

\begin{figure}[h!]
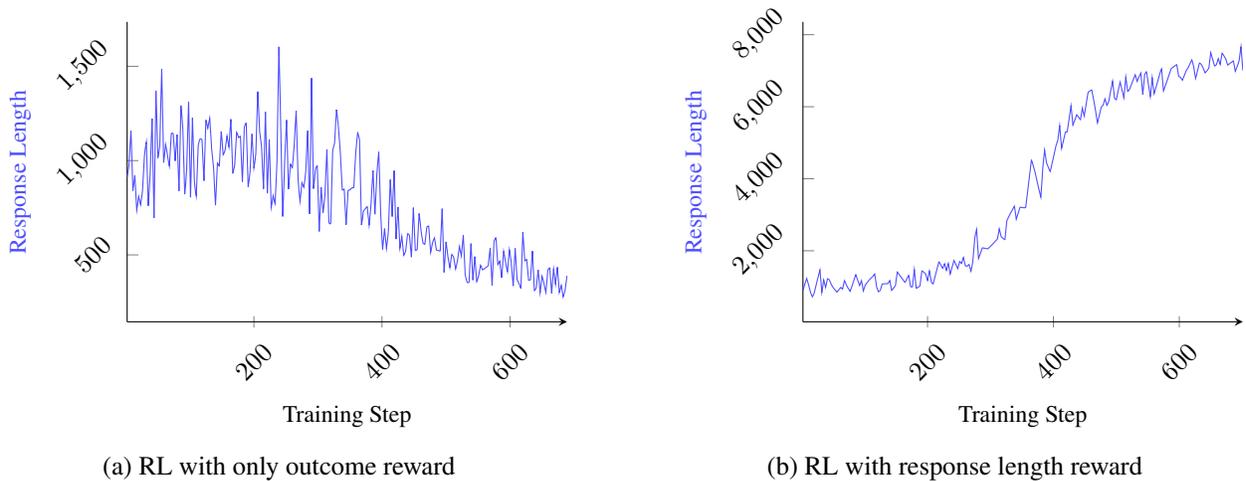

		\centering
		\begin{subfigure}[b]{0.45\textwidth}
			\centering
			\DualAxisPlot{figs/data/rl-code-qwen-2.5-7b_export.csv}{1}{0}{700}
			\caption{RL with only outcome reward}
            \label{fig:qwen_rl_code}
		\end{subfigure}
		\hfill
		\begin{subfigure}[b]{0.45\textwidth}
			\centering
			\DualAxisPlot{figs/data/rl-code-qwen-2.5-7b-w-length-bonus-v2_export.csv}{1}{20}{200}
			\caption{RL with response length reward}
            \label{fig:qwen_rl_code_with_len_p}
		\end{subfigure}
		\vspace{0.2em}
     \caption{Response length during RL training of Qwen2.5-7B-Instruct model in coding domain.}
     \label{fig:qwen_result_code}
	\end{figure}

We next ask if this behavior is specific to the math domain or the Qwen2.5-7B-Instruct model would also exhibit this behavior in other reasoning domains such as coding.
Our experiments in the coding domain are in \autoref{fig:qwen_result_code}.

We note that as opposed to the math domain, {\em in coding domain Qwen2.5-7B-Instruct model does not lead to search behavior with pure RL training using only outcome reward.}
On the other hand, as expected, reward with the response length leads to an increase on the search behavior. 
Thus, these ablation studies indicate that some form of explicit reward to encourage the thinking process may be more efficient and scalable way to enable search behavior in LLMs overall.

\section{Empirical Evaluation of RLSP}
\label{sec:empirical}
Having established that RL training even with simple exploration reward signal can lead to emergent search behaviors, in this section we evaluate the impact of full RLSP training on model performance within the math domain.
Our main goal in this section is to establish that search behavior leads to improved problem solving abilities.

Our experiments focus on two different model types with varying sizes: an 8B parameter model and a larger 32B parameter model.
Each model is trained and evaluated on a corresponding dataset tailored to its size, enabling a comprehensive analysis of the effectiveness of RLSP across different model scales.
The results are presented in two subsections, each dedicated to a specific model and dataset configuration.
The general experimental settings are provided in \autoref{sec:exp-settings}, while the specific hyperparameters for each experiment will be detailed in their respective sections.

We remark that our goal is not to compare against the SOTA performance of models on these benchmarks but {\em relative} improvement due to RLSP. 
Therefore, in all our experiments we compare the performance of same model trained using various strategies.
We did not optimize prompts and various other factors that can influence the performance of models both in base models and in RLSP training.
All our experiments are conducted on a cluster with 6 nodes (48 H100 GPUs), though most of our experiments utilize 24 H100 GPUs. 
Despite being performed at a small scale with only a few thousand training samples and iterations, our empirical evaluation provides strong evidence that RLSP can yield substantial gains when scaled up.

\subsection{Performance Analysis: 8B Model}
\label{sec:empirical_8b}

\subsubsection{Setup}
For this experiment, we use Llama-3.1-8B-Instruct \cite{llama-31-8b} as our base model.
As observed in \autoref{sec:can_pure_rl_lead_to_thinking}, while pure RL with an exploration reward signal encourages thinking behavior, the overall performance remains suboptimal due to the model's moderate size and limited capabilities.
Therefore, we conduct additional SFT on a math-specific dataset that includes long Chain-of-Thought (CoT) reasoning traces \cite{qwq-longcot}, sourced from the QwQ-32B-preview model \cite{qwq-32B-preview}.
To ensure the integrity of our evaluation, we decontaminate the dataset to remove any overlap with a wide range of test datasets.
Further details on the dataset curation process are provided in \autoref{sec:app-dataset-curation}.
Hyperparameters for the fine-tuning is reported in \autoref{sec:appendix_empirical_8b_sft_hyperparams}.

During the RL stage, the SFT model is trained with PPO on the training set of the MATH dataset \cite{hendrycks2021measuring}, utilizing only the (problem, answer) pairs. 
Problems from the training set that are already correctly solved by the SFT model are excluded. 
Details on the hyperparameters used for RL training are provided in \autoref{sec:appendix_empirical_8b_rl_hyperparams}.

For RLSP training, we use structured prompts with the GPT-4o model \cite{gpt-4o} to generate exploration reward scores. 
Further details about the prompts can be found in \autoref{sec:prompts}.

For evaluation, we measure the model's accuracy on the 500 test splits of the MATH dataset (a.k.a. MATH-500).

\subsubsection{Evaluation Results}
\autoref{tab:math_performance_8b} presents the results of this section.


\begin{table}[ht]
\caption{Performance comparison of RLSP training for MATH dataset. Base model is Llama-3.1-8B-Instruct. SFT model is the fine-tuned version of the base model with long Chain-of-Thought (CoT) reasoning traces. ER-RLSP represents RL training of the SFT model with our exploration reward. All results are based on pass@1 accuracy. The results demonstrate that RLSP training significantly boosts the performance of the base model.}
\label{tab:math_performance_8b}
\begin{center}
\begin{small}
\begin{sc}
\begin{tabular}{lcc}
\toprule
\textbf{Model}     & \textbf{MATH-500 (\%)}  \\ 
\midrule
Base          &        50.6    \\
SFT       &        70.2   \\
SFT + ER-RLSP    & 73.4        \\ 
\bottomrule
\end{tabular}
\end{sc}
\end{small}
\end{center}
\vskip -0.1in
\end{table}

We now analyze these findings and highlight key observations.
Firstly, SFT training yields a significant improvement in the base model's performance. 
This result underscores the importance of high-quality supervised demonstrations in enhancing the model’s reasoning capabilities, especially for small models.
Additionally, PPO training with our exploration reward yields a further 3\% improvement in performance. 
While this gain is not as pronounced as that achieved through SFT, it is expected, as SFT has already contributed substantial improvements, leaving limited room for additional gains through RL on the dataset MATH-500. 
Further, recall our (theory) discussion in \autoref{sec:theory} that long CoT SFT dataset itself can be thought of as constructed via a search procedure as it includes trajectories over rationales.
Thus, we conclude that even for small models, long CoT trajectories that incorporate search behaviors can lead to improved reasoning.

\subsection{Performance Analysis: 32B Model}
\label{sec:empirical_32b}

\subsubsection{Setup}
In this setup, we use Qwen2.5-32B-Instruct \cite{qwen-25-32b} as the base model.
Given its larger size and enhanced capabilities, we omit the SFT stage and focus on directly improving performance through RL training within our RLSP framework.

During the RL stage, the base model is trained using PPO on AIME 918 problems from the years 1983 to 2003.
We transition to the AIME dataset since the model already performs exceptionally well on the MATH dataset, nearing saturation. 
Thus, we focus on a more challenging setting to further assess and improve its capabilities.
Details on the hyperparameters used for RL training are provided in \autoref{sec:appendix_empirical_32b_rl_hyperparams}.

For evaluation, we assess the model's accuracy on AIME problems from the year 2024, as well as on the 500 test splits of the MATH dataset (referred to as MATH-500).

\subsubsection{Evaluation Results}
\autoref{tab:math_performance_32b} presents the results of this section.

\begin{table}[h]
\caption{Performance comparison of RLSP training for AIME and MATH datasets. Base model is Qwen2.5-32B-Instruct. ER-RLSP represents RL training of the base model with our exploration reward. All results are based on pass@1 accuracy. The results demonstrate that even with fewer than 1000 challenging math problems, RL training with our exploration reward can lead to significant performance improvements.}
\label{tab:math_performance_32b}
\begin{center}
\begin{small}
\begin{sc}
\begin{tabular}{lcc}
\toprule
\textbf{Model}     & \textbf{MATH (\%)} & \textbf{AIME (\%)} \\ 
\midrule
Base          & 81.6      & 13.3      \\
Base + ER-RLSP       &    83.2  &  23.3   \\ 
\bottomrule
\end{tabular}
\end{sc}
\end{small}
\end{center}

\end{table}

We highlight that even with a very small-scale set of challenging math problems as training data, our RL training approach can yield significant performance gains.
It is also noteworthy that while significant performance improvements are achieved on AIME 2024, we observe additional gains even on MATH-500, where the base model already demonstrates exceptional performance.

\subsection{Token Efficiency of RLSP vs Self-Consistency}
\label{sec:tokenefficiency}
A key intuition behind the RLSP framework is incentivizing the model to learn and apply search behavior during inference. 
We evaluate the token efficiency of the RLSP-trained model by asking: 
For a similar compute budget, what is the accuracy achieved by an RLSP-trained model versus a model using standard CoT with majority voting? 
We discuss the results in the following.

\autoref{tab:self_consistency_performance} presents the self-consistency performance of different models on the MATH-500 and AIME 2024 datasets, utilizing an average budget of 8192 tokens. 
Specifically, the Llama-3.1-8B-Instruct generates an average of 16 samples per test problem in MATH-500, while Qwen2.5-32B-Instruct produces an average of 8 samples per test problem in AIME 2024.
The self-consistency accuracy of Llama-3.1-8B-Instruct on MATH-500 is 61.8\%, while that of Qwen2.5-32B-Instruct on AIME24 is 20\%.
We note that these results fall short of our best performances, which reach 73.4\% (\autoref{tab:math_performance_8b}) and 23.3\% (\autoref{tab:math_performance_32b}), respectively.
Therefore, our study demonstrates that the RLSP framework enables better search and reasoning capabilities compared to majority voting or other independent sampling strategies.

\begin{table}[h]
\caption{Self-consistency performance metrics of Llama-3.1-8B-Instruct and Qwen2.5-32B-Instruct Models on MATH-500 and AIME 2024 with an average budget of 8192 tokens. Here \textbf{Avg Samples} = Average Samples per Problem, \textbf{SC Accuracy} = Self-Consistency Accuracy, and \textbf{Best Perf} = Best Performance.}
\label{tab:self_consistency_performance}
\begin{center}
\begin{small}
\begin{sc}
\begin{tabular}{lcc}
\toprule
\textbf{Metric} & \textbf{Llama-8B} & \textbf{Qwen-32B} \\
\midrule
\textbf{Benchmark} & MATH-500 & AIME 2024 \\
\textbf{Avg Samples} & 16 & 8 \\
\textbf{SC Accuracy} & 61.8\% & 20\% \\
\midrule
\textbf{SFT+ER-RLSP} & 73.4\% & 23.3\% \\
\bottomrule
\end{tabular}
\end{sc}
\end{small}
\end{center}
\vskip -0.1in
\end{table}


\section{Related Work}\label{sec:related-work}

In this section, we give a detailed literature survey.

\paragraph{Large Reasoning Models}
OpenAI’s o1 series \cite{openai-o1-mini, openai-o1, jaech2024openai} and other reasoning models \cite{gemini-2,deepseek-r1, qwq} represent a significant leap in reasoning capabilities, excelling in structured reasoning, systematic problem decomposition, and reliable handling of complex tasks. 
Testing is conducted on high-stakes benchmarks, including mathematics \cite{hendrycks2021measuring,gsm8k,aime24,amc23,math500}, competitive programming \cite{jain2024livecodebench,humaneval,codeforces}, and scientific problem-solving \cite{hendrycks2020measuring,rein2023gpqa}, often achieving performance levels that surpass human experts.

Recently, numerous open-source frameworks strive to replicate o1’s reasoning capabilities through diverse methodologies. At the post-training stage, frameworks such as \cite{wang2024openr, zhang2024rest, zhang2024llama} utilize automated data augmentation with MCTS, while \cite{huang2024o1, huang2025o1replicationjourney, skyt1, min2024imitate} exploit reasoning traces in long-CoT data. Process reward models are integrated into the training process by \cite{cui2024process, guan2025rstar}, boosting self-exploration. 
\cite{kimi-k15} introduces an effective RL framework emphasizing long-context scaling and robust policy optimization and incorporates with techniques for distilling long CoT reasoning into more efficient short-CoT models.
Meanwhile, \cite{deepseek-r1} highlights the potential of reinforcement learning in driving reasoning improvements on very large base models such as Deepseek V3 without process reward models or MCTS.
During inference, \cite{wang2024openr} employs majority-vote and beam search strategies, while \cite{zhang2024rest, qin2024o1, guan2025rstar, jiang2024technical} incorporate tree search-based inference. Additionally, \cite{zhang2024llama} emphasizes pairwise preference ranking and advanced exploration techniques, further enhancing inference performance.

\paragraph{Scaling Test-Time Compute}
Scaling test-time compute enhances reasoning capabilities by allocating more computational resources during inference. The test-time scaling laws demonstrate that increased deliberate reasoning (e.g., through additional token generation or iterative steps) directly improves accuracy, especially in complex tasks like mathematics, coding, and planning \cite{snell2024scaling,wu2024empirical,brown2024large,beeching2024scalingtesttimecompute}.
Recent work on simple test-time scaling \cite{muennighoff2025s1} shows that even small models can achieve significant improvements in reasoning tasks by strategically allocating inference budgets. In addition, RL-based scaling approaches \cite{hou2025advancing} show that inference scaling trends becomes more evident as training scales, reinforcing the connection between RL and test-time compute.

Various test-time search methods exploit this principle \cite{kang2024mindstar,wang2024q}. Majority vote aggregates predictions from multiple inference traces to refine accuracy. Tree search methods such as \cite{yao2024tree,hao2023reasoning,zhang2024accessing,qi2024mutual} systematically explore reasoning paths. Beam search \cite{smoke1961program} leverages the PRM and retains top-K paths at each step to improve sequence generation. Lookahead search \cite{snell2024scaling} further enhances exploration depth by evaluating paths using simulations. While these methods improve reasoning accuracy, they increase computational demand, highlighting the trade-off between performance and resource efficiency.

\paragraph{Math Data with Long Reasoning Traces}

Research indicates that combining System 2-inspired deliberate reasoning with System 1’s fast, heuristic-based thinking significantly enhances reasoning performance \cite{su2024dualformer}. 
To equip language models with System 2 reasoning techniques, recent advancements have explored approaches such as supervised fine-tuning (SFT) with extended chain-of-thought (CoT) reasoning \cite{bai2024longwriter,abdin2024phi,min2024imitate,huang2024o1,qin2024o1,wang2024enhancing,xu2024llava}. 
Training datasets with long reasoning traces, often distilled from advanced models like GPT \cite{li2024numinamath,luo2023wizardmath, yu2023metamath} and Qwen-qwq \cite{qwq-longcot}, are critical for fostering complex reasoning capabilities. 
To address the limitations of generator models, exploration-based techniques like MCTS \cite{guan2025rstar,zhang2024rest} and rejection sampling \cite{yuan2023scaling,brown2024large} systematically enhance reasoning by expanding decision paths. 
These methods enable language models to improve reasoning abilities and generate high-quality solutions to challenging problems, surpassing the constraints of their training data.

Recent studies further support the effectiveness of long CoT data. \cite{simplerl} demonstrates that reinforcement learning on long CoT-augmented data enables small models to exhibit emergent reasoning capabilities with significantly fewer examples. Similarly, \cite{open-r1} highlights that long CoT SFT leads to substantial improvements in problem-solving accuracy across mathematics and programming tasks. Furthermore, \cite{yeo2025demystifying} systematically investigates the factors driving long CoT emergence, revealing that while long CoT is not strictly necessary for reasoning ability, it substantially enhances training efficiency and stabilizes reinforcement learning by providing structured, verifiable reasoning trajectories. 
Meta-CoT \cite{xiang2025towards} provides a theoretical foundation for why long reasoning traces may enhance reasoning abilities. It argues that standard CoT fails to fully capture complex, non-linear thought processes and that explicitly modeling latent reasoning steps improves performance in high-difficulty tasks.

\paragraph{Self-Correction in Language Models}

Self-correction in LLMs has gained significant attention as a mechanism to enhance reasoning and problem-solving abilities. A range of techniques has been explored, spanning from fine-tuning methods to advanced reinforcement learning strategies. 
Fine-tuning approaches leverage curated data to train models for iterative corrections, improving their ability to refine responses \cite{min2024imitate, qu2024recursive, zhang2024small}. 
Prompt-based approaches focus on eliciting better outputs through iterative feedback loops \cite{huang2023large}, while inference strategies like MCTS are employed to refine reasoning by exploring diverse solution paths \cite{zhang2024accessing, yang2024qwen2, tian2024toward, wang2024towards}. 
Reinforcement learning methods such as Self-Correction via Reinforcement Learning (SCoRe) \cite{kumar2024training}, enhance self-correction through multi-turn RL using intrinsic rewards to guide learning.

While extrinsic feedback mechanisms remain effective, intrinsic self-correction—where models refine their outputs without external input—has proven challenging yet promising. Notable advancements include reward-based RL \cite{yang2024qwen2, choi2024self, kumar2024training} and curriculum preference learning \cite{wang2024towards}, which enable iterative refinement of reasoning steps. 
Unlike these approaches, we observe that self-correction behavior emerges naturally through unsupervised RL guided solely by a length penalty signals. This suggests that our method can be easily adapted to other domains without the need for domain-specific agent design or reward hacking.

\paragraph{Reinforcement Learning with Auxiliary rewards}

Reinforcement Learning (RL) with auxiliary rewards improves policy optimization by incorporating pseudo-reward signals that guide learning beyond task completion. 
The foundational work by Jaderberg et al. \cite{jaderberg2016reinforcement} uses auxiliary control and prediction tasks to accelerate learning. Recently, RL has been applied to enhance reasoning in large language models (LLMs). 
\cite{havrilla2024teaching} trained LLMs with outcome-based reward models (ORMs) and utilized a dense reward signal derived from comparing partial solutions to reference solutions.
\cite{setlur2024rewarding} introduced process advantage verifiers (PAVs), which assign stepwise rewards based on changes in correctness probability judged by a stronger prover policy.
\cite{luo2023wizardmath} employed process-supervised RL by distilling a process reward model (PRM) from GPT-4 annotations. In contrast, our work introduces a dense exploration reward that does not require a reference solution, making it more unsupervised and adaptable. Furthermore, rather than distilling PRM from GPT-4, we propose a more general approach to measuring creativity and reasoning effort, enhancing applicability across diverse domains.

\section{Conclusions, Limitations, and Future Work}
In this work we proposed a post-training technique called RLSP to enable thinking behavior for LRMs, showed promising results both in terms of performance and emergent behaviors.
More large scale experiments and analysis are necessary to fully understand capabilities and limitations of our work.
Needless to say, our work is a small step towards complex reasoning in LLMs and opens up several fascinating research directions:
How do we enable finer-grained test time search in LLMs where search time can have a direct influence on the quality of the solution so that model learns to differentiate between 1+1 = ? and the Riemann hypothesis.
What is the impact of context length on reasoning?
Can pure RL with no exploration reward lead to thinking behavior at some model scale? and what is the precise influence of pretraining data?
While all our models show interesting search behaviors such as backtracking and verification, none of those search strategies are surprising to us, and indirectly present in the pretraining data as  humans use those strategies as well. 
Is there a truly emergent behavior akin to "move 37'' that surpasses human reasoning or at least unexpected?
Finally, looking back at \autoref{fig:stagesofreasoing}, what other training recipes are needed to unlock even higher forms of reasoning such as creating abstractions and theories, and solving open ended problems like climate change or grand unified theory of everything.

\ifdefined\isarxiv
\bibliographystyle{alpha}
\bibliography{ref}
\else
\bibliography{ref}

\newcommand{\etalchar}[1]{$^{#1}$}
\begin{thebibliography}{HWW{\etalchar{+}}24}

\bibitem[AAB{\etalchar{+}}24]{abdin2024phi}
Marah Abdin, Jyoti Aneja, Harkirat Behl, S{\'e}bastien Bubeck, Ronen Eldan, Suriya Gunasekar, Michael Harrison, Russell~J Hewett, Mojan Javaheripi, Piero Kauffmann, et~al.
\newblock Phi-4 technical report.
\newblock {\em arXiv preprint arXiv:2412.08905}, 2024.

\bibitem[AM23]{amc23}
AI-MO.
\newblock Amc 2023.
\newblock \url{https://huggingface.co/datasets/AI-MO/aimo-validation\\ -amc}, 2023.

\bibitem[AM24a]{aime24}
AI-MO.
\newblock Aime 2024.
\newblock \url{https://huggingface.co/datasets/AI-MO/aimo-validation\\ -aime}, 2024.

\bibitem[AM24b]{numinamath-cot}
AI-MO.
\newblock Numinamath-cot.
\newblock \url{https://huggingface.co/datasets/AI-MO/NuminaMath-\\CoT}, 2024.

\bibitem[BJE{\etalchar{+}}24]{brown2024large}
Bradley Brown, Jordan Juravsky, Ryan Ehrlich, Ronald Clark, Quoc~V Le, Christopher R{\'e}, and Azalia Mirhoseini.
\newblock Large language monkeys: Scaling inference compute with repeated sampling.
\newblock {\em arXiv preprint arXiv:2407.21787}, 2024.

\bibitem[BKK{\etalchar{+}}22]{bai2022constitutional}
Yuntao Bai, Saurav Kadavath, Sandipan Kundu, Amanda Askell, Jackson Kernion, Andy Jones, Anna Chen, Anna Goldie, Azalia Mirhoseini, Cameron McKinnon, et~al.
\newblock Constitutional ai: Harmlessness from ai feedback.
\newblock {\em arXiv preprint arXiv:2212.08073}, 2022.

\bibitem[BTR24]{beeching2024scalingtesttimecompute}
Edward Beeching, Lewis Tunstall, and Sasha Rush.
\newblock Scaling test-time compute with open models, 2024.

\bibitem[BvWT25]{open-r1}
Elie Bakouch, Leandro von Werra, and Lewis Tunstall.
\newblock Open-r1: a fully open reproduction of deepseek-r1.
\newblock \url{https://huggingface.co/blog/open-r1}, 2025.

\bibitem[BZL{\etalchar{+}}24]{bai2024longwriter}
Yushi Bai, Jiajie Zhang, Xin Lv, Linzhi Zheng, Siqi Zhu, Lei Hou, Yuxiao Dong, Jie Tang, and Juanzi Li.
\newblock Longwriter: Unleashing 10,000+ word generation from long context llms.
\newblock {\em arXiv preprint arXiv:2408.07055}, 2024.

\bibitem[CAG{\etalchar{+}}24]{choi2024self}
Eugene Choi, Arash Ahmadian, Matthieu Geist, Oilvier Pietquin, and Mohammad~Gheshlaghi Azar.
\newblock Self-improving robust preference optimization.
\newblock {\em arXiv preprint arXiv:2406.01660}, 2024.

\bibitem[CLLF24]{chen2024alphamath}
Guoxin Chen, Minpeng Liao, Chengxi Li, and Kai Fan.
\newblock Alphamath almost zero: process supervision without process.
\newblock {\em arXiv preprint arXiv:2405.03553}, 2024.

\bibitem[Cod24]{codeforces}
CodeForces.
\newblock Codeforces dataset.
\newblock \url{https://codeforces.com/blog/entry/136853}, 2024.

\bibitem[CYW{\etalchar{+}}24]{cui2024process}
Ganqu Cui, Lifan Yuan, Zefan Wang, Hanbin Wang, Wendi Li, Bingxiang He, Yuchen Fan, Tianyu Yu, Qixin Xu, Weize Chen, Jiarui Yuan, Huayu Chen, Kaiyan Zhang, Xingtai Lv, Shuo Wang, Yuan Yao, Hao Peng, Yu~Cheng, Zhiyuan Liu, Maosong Sun, Bowen Zhou, and Ning Ding.
\newblock Process reinforcement through implicit rewards.
\newblock \url{https://curvy-check-498.notion.site/ Process-Reinforcement-through-\\ Implicit-Rewards-15f4fcb9c42180f1b49\\ 8cc9b2eaf896f}, 2024.
\newblock Notion Blog.

\bibitem[DA25]{deepseek-r1}
DeepSeek-AI.
\newblock Deepseek-r1: Incentivizing reasoning capability in llms via reinforcement learning.
\newblock \url{https://github.com/deepseek-ai/DeepSeek-R1/blob/main/DeepSeek_R1.pdf}, 2025.

\bibitem[Dao24]{dao2023flashattention2}
Tri Dao.
\newblock Flash{A}ttention-2: Faster attention with better parallelism and work partitioning.
\newblock In {\em International Conference on Learning Representations (ICLR)}, 2024.

\bibitem[Dee24]{alphaproof}
Google Deepmind.
\newblock Ai achieves silver-medal standard solving international mathematical olympiad problems, 2024.

\bibitem[FWW{\etalchar{+}}23]{feng2023alphazero}
Xidong Feng, Ziyu Wan, Muning Wen, Stephen~Marcus McAleer, Ying Wen, Weinan Zhang, and Jun Wang.
\newblock Alphazero-like tree-search can guide large language model decoding and training.
\newblock {\em arXiv preprint arXiv:2309.17179}, 2023.

\bibitem[Goo24]{gemini-2}
Google.
\newblock Introducing gemini 2.0: our new ai model for the agentic era.
\newblock \url{https://blog.google/technology/google-deepmind/google-gemini-ai-update-december-2024/#ceo-message}, 2024.

\bibitem[GZL{\etalchar{+}}25]{guan2025rstar}
Xinyu Guan, Li~Lyna Zhang, Yifei Liu, Ning Shang, Youran Sun, Yi~Zhu, Fan Yang, and Mao Yang.
\newblock rstar-math: Small llms can master math reasoning with self-evolved deep thinking.
\newblock {\em arXiv preprint arXiv:2501.04519}, 2025.

\bibitem[HBB{\etalchar{+}}20]{hendrycks2020measuring}
Dan Hendrycks, Collin Burns, Steven Basart, Andy Zou, Mantas Mazeika, Dawn Song, and Jacob Steinhardt.
\newblock Measuring massive multitask language understanding.
\newblock {\em arXiv preprint arXiv:2009.03300}, 2020.

\bibitem[HBK{\etalchar{+}}21]{hendrycks2021measuring}
Dan Hendrycks, Collin Burns, Saurav Kadavath, Akul Arora, Steven Basart, Eric Tang, Dawn Song, and Jacob Steinhardt.
\newblock Measuring mathematical problem solving with the math dataset.
\newblock {\em arXiv preprint arXiv:2103.03874}, 2021.

\bibitem[HCM{\etalchar{+}}23]{huang2023large}
Jie Huang, Xinyun Chen, Swaroop Mishra, Huaixiu~Steven Zheng, Adams~Wei Yu, Xinying Song, and Denny Zhou.
\newblock Large language models cannot self-correct reasoning yet.
\newblock {\em arXiv preprint arXiv:2310.01798}, 2023.

\bibitem[HDR{\etalchar{+}}24]{havrilla2024teaching}
Alex Havrilla, Yuqing Du, Sharath~Chandra Raparthy, Christoforos Nalmpantis, Jane Dwivedi-Yu, Maksym Zhuravinskyi, Eric Hambro, Sainbayar Sukhbaatar, and Roberta Raileanu.
\newblock Teaching large language models to reason with reinforcement learning.
\newblock {\em arXiv preprint arXiv:2403.04642}, 2024.

\bibitem[HGH{\etalchar{+}}25]{huang2025o1replicationjourney}
Zhongzhen Huang, Gui Geng, Shengyi Hua, Zhen Huang, Haoyang Zou, Shaoting Zhang, Pengfei Liu, and Xiaofan Zhang.
\newblock O1 replication journey -- part 3: Inference-time scaling for medical reasoning.
\newblock {\em arXiv preprint arXiv:2501.06458}, 2025.

\bibitem[HGM{\etalchar{+}}23]{hao2023reasoning}
Shibo Hao, Yi~Gu, Haodi Ma, Joshua~Jiahua Hong, Zhen Wang, Daisy~Zhe Wang, and Zhiting Hu.
\newblock Reasoning with language model is planning with world model.
\newblock {\em arXiv preprint arXiv:2305.14992}, 2023.

\bibitem[HLG{\etalchar{+}}24]{gpt-4o}
Aaron Hurst, Adam Lerer, Adam~P Goucher, Adam Perelman, Aditya Ramesh, Aidan Clark, AJ~Ostrow, Akila Welihinda, Alan Hayes, Alec Radford, et~al.
\newblock Gpt-4o system card.
\newblock {\em arXiv preprint arXiv:2410.21276}, 2024.

\bibitem[HLL{\etalchar{+}}25]{hou2025advancing}
Zhenyu Hou, Xin Lv, Rui Lu, Jiajie Zhang, Yujiang Li, Zijun Yao, Juanzi Li, Jie Tang, and Yuxiao Dong.
\newblock Advancing language model reasoning through reinforcement learning and inference scaling.
\newblock {\em arXiv preprint arXiv:2501.11651}, 2025.

\bibitem[HWW{\etalchar{+}}24]{hu2024openrlhf}
Jian Hu, Xibin Wu, Weixun Wang, Dehao Zhang, Yu~Cao, et~al.
\newblock Openrlhf: An easy-to-use, scalable and high-performance rlhf framework.
\newblock {\em arXiv preprint arXiv:2405.11143}, 2024.

\bibitem[HZL{\etalchar{+}}24]{huang2024o1}
Zhen Huang, Haoyang Zou, Xuefeng Li, Yixiu Liu, Yuxiang Zheng, Ethan Chern, Shijie Xia, Yiwei Qin, Weizhe Yuan, and Pengfei Liu.
\newblock O1 replication journey--part 2: Surpassing o1-preview through simple distillation, big progress or bitter lesson?
\newblock {\em arXiv preprint arXiv:2411.16489}, 2024.

\bibitem[JCM{\etalchar{+}}24]{jiang2024technical}
Jinhao Jiang, Zhipeng Chen, Yingqian Min, Jie Chen, Xiaoxue Cheng, Jiapeng Wang, Yiru Tang, Haoxiang Sun, Jia Deng, Wayne~Xin Zhao, et~al.
\newblock Technical report: Enhancing llm reasoning with reward-guided tree search.
\newblock {\em arXiv preprint arXiv:2411.11694}, 2024.

\bibitem[JHG{\etalchar{+}}24]{jain2024livecodebench}
Naman Jain, King Han, Alex Gu, Wen-Ding Li, Fanjia Yan, Tianjun Zhang, Sida Wang, Armando Solar-Lezama, Koushik Sen, and Ion Stoica.
\newblock Livecodebench: Holistic and contamination free evaluation of large language models for code.
\newblock {\em arXiv preprint arXiv:2403.07974}, 2024.

\bibitem[JKL{\etalchar{+}}24]{jaech2024openai}
Aaron Jaech, Adam Kalai, Adam Lerer, Adam Richardson, Ahmed El-Kishky, Aiden Low, Alec Helyar, Aleksander Madry, Alex Beutel, Alex Carney, et~al.
\newblock Openai o1 system card.
\newblock {\em arXiv preprint arXiv:2412.16720}, 2024.

\bibitem[JMC{\etalchar{+}}16]{jaderberg2016reinforcement}
Max Jaderberg, Volodymyr Mnih, Wojciech~Marian Czarnecki, Tom Schaul, Joel~Z Leibo, David Silver, and Koray Kavukcuoglu.
\newblock Reinforcement learning with unsupervised auxiliary tasks.
\newblock {\em arXiv preprint arXiv:1611.05397}, 2016.

\bibitem[Kim25]{kimi-k15}
Kimi.
\newblock Kimi k1.5: Scaling reinforcement learning with llms.
\newblock \url{https://github.com/MoonshotAI/Kimi-k1.5/blob/main/Kimi_k1.5.pdf}, 2025.

\bibitem[KLC{\etalchar{+}}24]{kang2024mindstar}
Jikun Kang, Xin~Zhe Li, Xi~Chen, Amirreza Kazemi, Qianyi Sun, Boxing Chen, Dong Li, Xu~He, Quan He, Feng Wen, et~al.
\newblock Mindstar: Enhancing math reasoning in pre-trained llms at inference time.
\newblock {\em arXiv preprint arXiv:2405.16265}, 2024.

\bibitem[KZA{\etalchar{+}}24]{kumar2024training}
Aviral Kumar, Vincent Zhuang, Rishabh Agarwal, Yi~Su, John~D Co-Reyes, Avi Singh, Kate Baumli, Shariq Iqbal, Colton Bishop, Rebecca Roelofs, et~al.
\newblock Training language models to self-correct via reinforcement learning.
\newblock {\em arXiv preprint arXiv:2409.12917}, 2024.

\bibitem[LBT{\etalchar{+}}24]{li2024numinamath}
Jia Li, Edward Beeching, Lewis Tunstall, Ben Lipkin, Roman Soletskyi, Shengyi Huang, Kashif Rasul, Longhui Yu, Albert~Q Jiang, Ziju Shen, et~al.
\newblock Numinamath: The largest public dataset in ai4maths with 860k pairs of competition math problems and solutions.
\newblock {\em Hugging Face repository}, 13, 2024.

\bibitem[LKB{\etalchar{+}}23]{lightman2023let}
Hunter Lightman, Vineet Kosaraju, Yura Burda, Harri Edwards, Bowen Baker, Teddy Lee, Jan Leike, John Schulman, Ilya Sutskever, and Karl Cobbe.
\newblock Let's verify step by step.
\newblock {\em arXiv preprint arXiv:2305.20050}, 2023.

\bibitem[LLZM24]{li2024chain}
Zhiyuan Li, Hong Liu, Denny Zhou, and Tengyu Ma.
\newblock Chain of thought empowers transformers to solve inherently serial problems.
\newblock {\em arXiv preprint arXiv:2402.12875}, 2024.

\bibitem[LSX{\etalchar{+}}23]{luo2023wizardmath}
Haipeng Luo, Qingfeng Sun, Can Xu, Pu~Zhao, Jianguang Lou, Chongyang Tao, Xiubo Geng, Qingwei Lin, Shifeng Chen, and Dongmei Zhang.
\newblock Wizardmath: Empowering mathematical reasoning for large language models via reinforced evol-instruct.
\newblock {\em arXiv preprint arXiv:2308.09583}, 2023.

\bibitem[LYF{\etalchar{+}}23]{liu2023pre}
Pengfei Liu, Weizhe Yuan, Jinlan Fu, Zhengbao Jiang, Hiroaki Hayashi, and Graham Neubig.
\newblock Pre-train, prompt, and predict: A systematic survey of prompting methods in natural language processing.
\newblock {\em ACM Computing Surveys}, 55(9):1--35, 2023.

\bibitem[MCJ{\etalchar{+}}24]{min2024imitate}
Yingqian Min, Zhipeng Chen, Jinhao Jiang, Jie Chen, Jia Deng, Yiwen Hu, Yiru Tang, Jiapeng Wang, Xiaoxue Cheng, Huatong Song, et~al.
\newblock Imitate, explore, and self-improve: A reproduction report on slow-thinking reasoning systems.
\newblock {\em arXiv preprint arXiv:2412.09413}, 2024.

\bibitem[Met24]{llama-31-8b}
Meta.
\newblock Llama-3.1-8b.
\newblock \url{https://huggingface.co/meta-llama/Llama-3.1-8B}, 2024.

\bibitem[MNW{\etalchar{+}}18]{moritz2018ray}
Philipp Moritz, Robert Nishihara, Stephanie Wang, Alexey Tumanov, Richard Liaw, Eric Liang, Melih Elibol, Zongheng Yang, William Paul, Michael~I Jordan, et~al.
\newblock Ray: A distributed framework for emerging $\{$AI$\}$ applications.
\newblock In {\em 13th USENIX symposium on operating systems design and implementation (OSDI 18)}, pages 561--577, 2018.

\bibitem[MS23]{merrill2023expresssive}
William Merrill and Ashish Sabharwal.
\newblock The expresssive power of transformers with chain of thought.
\newblock {\em arXiv preprint arXiv:2310.07923}, 2023.

\bibitem[MYS{\etalchar{+}}25]{muennighoff2025s1}
Niklas Muennighoff, Zitong Yang, Weijia Shi, Xiang~Lisa Li, Li~Fei-Fei, Hannaneh Hajishirzi, Luke Zettlemoyer, Percy Liang, Emmanuel Cand{\`e}s, and Tatsunori Hashimoto.
\newblock s1: Simple test-time scaling.
\newblock {\em arXiv preprint arXiv:2501.19393}, 2025.

\bibitem[Nov25]{skyt1}
NovaSky.
\newblock Sky-t1: Train your own o1 preview model within \$450.
\newblock \url{https://novasky-ai.github.io/posts/sky-t1/}, 2025.

\bibitem[Ope22]{gsm8k}
OpenAI.
\newblock Gsm8k.
\newblock \url{https://huggingface.co/datasets/openai/gsm8k}, 2022.

\bibitem[Ope24a]{openai-o1}
OpenAI.
\newblock Learning to reason with llms.
\newblock \url{https://openai.com/index/learning-to-\\ reason-with-llms/}, 2024.

\bibitem[Ope24b]{math500}
OpenAI.
\newblock Math-500.
\newblock \url{https://huggingface.co/datasets/HuggingFaceH4/MATH-500}, 2024.

\bibitem[Ope24c]{humaneval}
OpenAI.
\newblock Openai humaneval.
\newblock \url{https://huggingface.co/datasets/openai/openai_humaneval}, 2024.

\bibitem[Ope24d]{openai-o1-mini}
OpenAI.
\newblock Openai o1-mini advancing cost-efficient reasoning.
\newblock \url{https://openai.com/index/openai-o1-mini-advancing-cost-\\ efficient-reasoning/}, 2024.

\bibitem[OWJ{\etalchar{+}}22]{ouyang2022training}
Long Ouyang, Jeffrey Wu, Xu~Jiang, Diogo Almeida, Carroll Wainwright, Pamela Mishkin, Chong Zhang, Sandhini Agarwal, Katarina Slama, Alex Ray, et~al.
\newblock Training language models to follow instructions with human feedback.
\newblock {\em Advances in neural information processing systems}, 35:27730--27744, 2022.

\bibitem[QLZ{\etalchar{+}}24]{qin2024o1}
Yiwei Qin, Xuefeng Li, Haoyang Zou, Yixiu Liu, Shijie Xia, Zhen Huang, Yixin Ye, Weizhe Yuan, Hector Liu, Yuanzhi Li, et~al.
\newblock O1 replication journey: A strategic progress report--part 1.
\newblock {\em arXiv preprint arXiv:2410.18982}, 2024.

\bibitem[QMX{\etalchar{+}}24]{qi2024mutual}
Zhenting Qi, Mingyuan Ma, Jiahang Xu, Li~Lyna Zhang, Fan Yang, and Mao Yang.
\newblock Mutual reasoning makes smaller llms stronger problem-solvers.
\newblock {\em arXiv preprint arXiv:2408.06195}, 2024.

\bibitem[Qwe24a]{qwen-25-32b}
Qwen.
\newblock Qwen-2.5-32b.
\newblock \url{https://huggingface.co/Qwen/Qwen2.5-32B}, 2024.

\bibitem[Qwe24b]{qwq-32B-preview}
Qwen.
\newblock Qwq-32b-preview.
\newblock \url{https://huggingface.co/Qwen/QwQ-32B-Preview}, 2024.

\bibitem[Qwe24c]{qwq-longcot}
Qwen.
\newblock Qwq-longcot-130k-cleaned.
\newblock \url{https://huggingface.co/datasets/gghfez/QwQ-LongCoT-130K-cleaned}, 2024.

\bibitem[Qwe24d]{qwq}
Qwen.
\newblock Qwq: Reflect deeply on the boundaries of the unknown.
\newblock \url{https://qwenlm.github.io/blog/qwq-32b-preview}, 2024.

\bibitem[QZGK24]{qu2024recursive}
Yuxiao Qu, Tianjun Zhang, Naman Garg, and Aviral Kumar.
\newblock Recursive introspection: Teaching language model agents how to self-improve.
\newblock {\em arXiv preprint arXiv:2407.18219}, 2024.

\bibitem[RHS{\etalchar{+}}23]{rein2023gpqa}
David Rein, Betty~Li Hou, Asa~Cooper Stickland, Jackson Petty, Richard~Yuanzhe Pang, Julien Dirani, Julian Michael, and Samuel~R Bowman.
\newblock Gpqa: A graduate-level google-proof q\&a benchmark.
\newblock {\em arXiv preprint arXiv:2311.12022}, 2023.

\bibitem[SD61]{smoke1961program}
W~Smoke and E~Dubinsky.
\newblock A program for the machine translation of natural languages.
\newblock {\em Mech. Transl. Comput. Linguistics}, 6:2--10, 1961.

\bibitem[SLXK24]{snell2024scaling}
Charlie Snell, Jaehoon Lee, Kelvin Xu, and Aviral Kumar.
\newblock Scaling llm test-time compute optimally can be more effective than scaling model parameters.
\newblock {\em arXiv preprint arXiv:2408.03314}, 2024.

\bibitem[SNF{\etalchar{+}}24]{setlur2024rewarding}
Amrith Setlur, Chirag Nagpal, Adam Fisch, Xinyang Geng, Jacob Eisenstein, Rishabh Agarwal, Alekh Agarwal, Jonathan Berant, and Aviral Kumar.
\newblock Rewarding progress: Scaling automated process verifiers for llm reasoning.
\newblock {\em arXiv preprint arXiv:2410.08146}, 2024.

\bibitem[SSR{\etalchar{+}}24]{su2024dualformer}
DiJia Su, Sainbayar Sukhbaatar, Michael Rabbat, Yuandong Tian, and Qinqing Zheng.
\newblock Dualformer: Controllable fast and slow thinking by learning with randomized reasoning traces.
\newblock {\em arXiv preprint arXiv:2410.09918}, 2024.

\bibitem[SWD{\etalchar{+}}17]{schulman2017proximal}
John Schulman, Filip Wolski, Prafulla Dhariwal, Alec Radford, and Oleg Klimov.
\newblock Proximal policy optimization algorithms.
\newblock {\em arXiv preprint arXiv:1707.06347}, 2017.

\bibitem[TPS{\etalchar{+}}24]{tian2024toward}
Ye~Tian, Baolin Peng, Linfeng Song, Lifeng Jin, Dian Yu, Haitao Mi, and Dong Yu.
\newblock Toward self-improvement of llms via imagination, searching, and criticizing.
\newblock {\em arXiv preprint arXiv:2404.12253}, 2024.

\bibitem[TWL{\etalchar{+}}24]{trinh2024alphageometry}
Trieu~H Trinh, Yuhuai Wu, Quoc~V Le, He~He, and Thang Luong.
\newblock Solving olympiad geometry without human demonstrations.
\newblock {\em Nature}, 625(7995):476--482, 2024.

\bibitem[UKK{\etalchar{+}}22]{uesato2022solving}
Jonathan Uesato, Nate Kushman, Ramana Kumar, Francis Song, Noah Siegel, Lisa Wang, Antonia Creswell, Geoffrey Irving, and Irina Higgins.
\newblock Solving math word problems with process-and outcome-based feedback.
\newblock {\em arXiv preprint arXiv:2211.14275}, 2022.

\bibitem[WCW{\etalchar{+}}24]{wang2024enhancing}
Weiyun Wang, Zhe Chen, Wenhai Wang, Yue Cao, Yangzhou Liu, Zhangwei Gao, Jinguo Zhu, Xizhou Zhu, Lewei Lu, Yu~Qiao, et~al.
\newblock Enhancing the reasoning ability of multimodal large language models via mixed preference optimization.
\newblock {\em arXiv preprint arXiv:2411.10442}, 2024.

\bibitem[WDL{\etalchar{+}}24]{wang2024q}
Chaojie Wang, Yanchen Deng, Zhiyi Lyu, Liang Zeng, Jujie He, Shuicheng Yan, and Bo~An.
\newblock Q*: Improving multi-step reasoning for llms with deliberative planning.
\newblock {\em arXiv preprint arXiv:2406.14283}, 2024.

\bibitem[WFW{\etalchar{+}}24]{wang2024openr}
Jun Wang, Meng Fang, Ziyu Wan, Muning Wen, Jiachen Zhu, Anjie Liu, Ziqin Gong, Yan Song, Lei Chen, Lionel~M Ni, et~al.
\newblock Openr: An open source framework for advanced reasoning with large language models.
\newblock {\em arXiv preprint arXiv:2410.09671}, 2024.

\bibitem[WSL{\etalchar{+}}24]{wu2024empirical}
Yangzhen Wu, Zhiqing Sun, Shanda Li, Sean Welleck, and Yiming Yang.
\newblock An empirical analysis of compute-optimal inference for problem-solving with language models.
\newblock {\em Under review}, 2024.

\bibitem[WST{\etalchar{+}}24]{wang2024towards}
Xiyao Wang, Linfeng Song, Ye~Tian, Dian Yu, Baolin Peng, Haitao Mi, Furong Huang, and Dong Yu.
\newblock Towards self-improvement of llms via mcts: Leveraging stepwise knowledge with curriculum preference learning.
\newblock {\em arXiv preprint arXiv:2410.06508}, 2024.

\bibitem[WWS{\etalchar{+}}22]{wang2022self}
Xuezhi Wang, Jason Wei, Dale Schuurmans, Quoc Le, Ed~Chi, Sharan Narang, Aakanksha Chowdhery, and Denny Zhou.
\newblock Self-consistency improves chain of thought reasoning in language models.
\newblock {\em arXiv preprint arXiv:2203.11171}, 2022.

\bibitem[XJH{\etalchar{+}}24]{xu2024llava}
Guowei Xu, Peng Jin, Li~Hao, Yibing Song, Lichao Sun, and Li~Yuan.
\newblock Llava-o1: Let vision language models reason step-by-step.
\newblock {\em arXiv preprint arXiv:2411.10440}, 2024.

\bibitem[XSG{\etalchar{+}}25]{xiang2025towards}
Violet Xiang, Charlie Snell, Kanishk Gandhi, Alon Albalak, Anikait Singh, Chase Blagden, Duy Phung, Rafael Rafailov, Nathan Lile, Dakota Mahan, et~al.
\newblock Towards system 2 reasoning in llms: Learning how to think with meta chain-of-though.
\newblock {\em arXiv preprint arXiv:2501.04682}, 2025.

\bibitem[YJS{\etalchar{+}}23]{yu2023metamath}
Longhui Yu, Weisen Jiang, Han Shi, Jincheng Yu, Zhengying Liu, Yu~Zhang, James~T Kwok, Zhenguo Li, Adrian Weller, and Weiyang Liu.
\newblock Metamath: Bootstrap your own mathematical questions for large language models.
\newblock {\em arXiv preprint arXiv:2309.12284}, 2023.

\bibitem[YTN{\etalchar{+}}25]{yeo2025demystifying}
Edward Yeo, Yuxuan Tong, Morry Niu, Graham Neubig, and Xiang Yue.
\newblock Demystifying long chain-of-thought reasoning in llms.
\newblock {\em arXiv preprint arXiv:2502.03373}, 2025.

\bibitem[YYL{\etalchar{+}}23]{yuan2023scaling}
Zheng Yuan, Hongyi Yuan, Chengpeng Li, Guanting Dong, Keming Lu, Chuanqi Tan, Chang Zhou, and Jingren Zhou.
\newblock Scaling relationship on learning mathematical reasoning with large language models.
\newblock {\em arXiv preprint arXiv:2308.01825}, 2023.

\bibitem[YYZ{\etalchar{+}}24]{yao2024tree}
Shunyu Yao, Dian Yu, Jeffrey Zhao, Izhak Shafran, Tom Griffiths, Yuan Cao, and Karthik Narasimhan.
\newblock Tree of thoughts: Deliberate problem solving with large language models.
\newblock {\em Advances in Neural Information Processing Systems}, 36, 2024.

\bibitem[YZH{\etalchar{+}}24]{yang2024qwen2}
An~Yang, Beichen Zhang, Binyuan Hui, Bofei Gao, Bowen Yu, Chengpeng Li, Dayiheng Liu, Jianhong Tu, Jingren Zhou, Junyang Lin, et~al.
\newblock Qwen2. 5-math technical report: Toward mathematical expert model via self-improvement.
\newblock {\em arXiv preprint arXiv:2409.12122}, 2024.

\bibitem[ZHL{\etalchar{+}}25]{simplerl}
Weihao Zeng, Yuzhen Huang, Wei Liu, Keqing He, Qian Liu, Zejun Ma, and Junxian He.
\newblock 7b model and 8k examples: Emerging reasoning with reinforcement learning is both effective and efficient.
\newblock \url{https://hkust-nlp.notion.site/simplerl-reason#18439bdc1c6b8083ba31f9cc912cf7f0}, 2025.

\bibitem[ZKL{\etalchar{+}}24]{zhang2024small}
Yunxiang Zhang, Muhammad Khalifa, Lajanugen Logeswaran, Jaekyeom Kim, Moontae Lee, Honglak Lee, and Lu~Wang.
\newblock Small language models need strong verifiers to self-correct reasoning.
\newblock {\em arXiv preprint arXiv:2404.17140}, 2024.

\bibitem[ZLH{\etalchar{+}}24]{zhang2024accessing}
Di~Zhang, Jiatong Li, Xiaoshui Huang, Dongzhan Zhou, Yuqiang Li, and Wanli Ouyang.
\newblock Accessing gpt-4 level mathematical olympiad solutions via monte carlo tree self-refine with llama-3 8b.
\newblock {\em arXiv preprint arXiv:2406.07394}, 2024.

\bibitem[ZWL{\etalchar{+}}24]{zhang2024llama}
Di~Zhang, Jianbo Wu, Jingdi Lei, Tong Che, Jiatong Li, Tong Xie, Xiaoshui Huang, Shufei Zhang, Marco Pavone, Yuqiang Li, et~al.
\newblock Llama-berry: Pairwise optimization for o1-like olympiad-level mathematical reasoning.
\newblock {\em arXiv preprint arXiv:2410.02884}, 2024.

\bibitem[ZZH{\etalchar{+}}24]{zhang2024rest}
Dan Zhang, Sining Zhoubian, Ziniu Hu, Yisong Yue, Yuxiao Dong, and Jie Tang.
\newblock Rest-mcts*: Llm self-training via process reward guided tree search.
\newblock {\em arXiv preprint arXiv:2406.03816}, 2024.

\end{thebibliography}
\bibliographystyle{alpha}

\fi

\newpage
\onecolumn
\appendix


\section{Examples of RLSP Emergence Behavior}\label{sec:app-rlsp-emergence}

In this section, we present three examples demonstrating the emergence of self-verification, backtracking, and self-correction behaviors. The models are trained solely using PPO with a length penalty reward on the Llama-3.1-8B-Instruct model, without any supervised fine-tuning. All steps exhibiting these behaviors are highlighted in red.

\begin{itemize}
    \item \textbf{Self-Verification:} The model explicitly checks its own reasoning steps to confirm correctness before finalizing an answer. This often involves re-evaluating computations or validating logical consistency.  
    \item \textbf{Consideration of alternative possibilities:} The model searches over multiple rationales and explores alternative approaches and recognizes inconsistencies or similarities. 
    \item \textbf{Self-Correction:} The model identifies errors in its intermediate steps and revises its solution in the subsequent steps.
    \item \textbf{Backtracking:} Upon identifying errors or dead ends, the model goes back to an earlier step and considers a different reasoning path. 
\end{itemize}

These examples provide strong evidence that RL facilitates the emergence of more systematic and reflective problem-solving behaviours without explicit supervision.

\vskip 2em

\begin{promptbox}[box:fig1-full]{Full Self-Verification Behavior for \autoref{fig:emergence_example}}

\textbf{Question:} The positive three-digit integer $N$ has a ones digit of $0$. What is the probability that $N$ is divisible by $4$? Express your answer as a common fraction.

\textbf{Step 1}: Understand the conditions for divisibility by 4

A number is divisible by 4 if the number formed by its last two digits is divisible by 4.
 
\textbf{Step 2}: Recognize the possible last two digits for N

Since N is a three-digit number and its ones digit is 0, we only need to consider the last two digits which can range from 00 to 90, in increments of 10.
 
\textbf{Step 3}: List the possible last two digits of N

The possible last two digits are 00, 10, 20, 30, 40, 50, 60, 70, 80, and 90.
 
\textbf{Step 4}: Determine how many of these are divisible by 4

Among 00, 10, 20, 30, 40, 50, 60, 70, 80, and 90, the numbers that are divisible by 4 are 00, 20, 40, 60, and 80.
 
\textbf{Step 5}: Count the total number of possibilities

There are 10 possibilities in total.
 
\textbf{Step 6}: Calculate the probability that N is divisible by 4

The probability that N is divisible by 4 is the number of successful outcomes (numbers divisible by 4) divided by the total number of outcomes.
 
\textbf{Step 7}: Calculate the probability

There are 5 successful outcomes (00, 20, 40, 60, 80) and 10 total outcomes. So the probability is 5/10 which simplifies to 1/2.
 
\textbf{Step 8}: Consider the divisibility rule for 4

To be divisible by 4, the last two digits of N must be divisible by 4. This means for every 10 numbers from 00 to 90, exactly one will be divisible by 4.
 
{\color{red}\textbf{Step 9}: Verify the previous conclusion}

Since we know that the numbers 00, 20, 40, 60, and 80 are divisible by 4 and these are evenly spaced, we can confirm that there is one number out of every ten from 00 to 90 that is divisible by 4.
 
\textbf{Step 10}: Calculate the probability

The probability that N is divisible by 4 is therefore 1/2, since half the numbers from 00 to 90 are divisible by 4.
 
\textbf{Step 11}: Consider the fact that 00 is a special case

In a set of numbers from 00 to 90, 00 is counted as 1 of the numbers, and it is divisible by 4.
 
\textbf{Step 12}: Consider the implication of the 00 case for the probability

Since 00 is divisible by 4, it does not change the fact that half the numbers from 00 to 90 are divisible by 4.
 
\textbf{Step 13}: Consider the fact that there are 10 numbers in the set from 00 to 90

This confirms that 5 of the numbers are divisible by 4 and 5 are not.
 
\textbf{Step 14}: Calculate the probability that N is divisible by 4

The probability that N is divisible by 4 is therefore 5/10, which simplifies to 1/2.
 
\textbf{Step 15}: Understand that we don't need to calculate the probability for all 3-digit numbers

Since we are only concerned with three-digit numbers that have a ones digit of 0, the set of numbers we are considering is much smaller than all three-digit numbers.
 
\textbf{Step 16}: Realize that the probability that N is divisible by 4 is still 1/2

This is because 1/2 of the numbers from 00 to 90 are divisible by 4, regardless of the other digits in the number.
 
\textbf{Step 17}: Consider the fact that the hundreds digit does not affect the probability

Since the hundreds digit does not affect whether a number is divisible by 4, it does not change the probability.
 
\textbf{Step 18}: Consider the fact that the tens digit does not affect the probability

Similarly, the tens digit does not affect whether a number is divisible by 4.
 
\textbf{Step 19}: Realize that the probability that N is divisible by 4 is determined by the last two digits

Since the ones digit is always 0, we only need to consider the last two digits.
 
{\color{red}\textbf{Step 20}: Understand that this confirms that 1/2 of the numbers are divisible by 4}

Since there are 5 numbers from 00 to 90 that are divisible by 4, and 10 total numbers, the probability that N is divisible by 4 is 5/10, which simplifies to 1/2.
 
\textbf{Step 21}: Consider the fact that N is a three-digit number with a ones digit of 0

This means that the last two digits can range from 00 to 90.
 
{\color{red} \textbf{Step 22}: Understand that this confirms our previous conclusion}

We have already determined that 5 of the numbers from 00 to 90 are divisible by 4.
 
\textbf{Step 23}: Realize that the probability that N is divisible by 4 is 1/2

Since 5 of the 10 numbers from 00 to 90 are divisible by 4, the probability is 5/10, which simplifies to 1/2.

The final answer is: $\boxed{\frac{1}{2}}$
\end{promptbox}

\vskip 2em

\begin{promptbox}{Example of Self-Verification Behavior}
Problem: The equation of the line that passes through the points $(-3,5)$ and $(0,-4)$ can be expressed in the form $y=mx+b$. What is the value of $m+b$?

assistant 

\textbf{\#\# Step 1:}
To find the equation of the line that passes through the points $(-3,5)$ and $(0,-4)$, we should use the slope-intercept form, which is $y=mx+b$, where $m$ is the slope of the line and $b$ is the y-intercept.
 
\textbf{\#\# Step 2:}
The slope $m$ of a line passing through two points $(x_1,y_1)$ and $(x_2,y_2)$ can be calculated using the formula: $m=\dfrac{y_2-y_1}{x_2-x_1}$. Let's use the given points to find the slope $m$.
 
\textbf{\#\# Step 3:}
Plugging the given points into the slope formula: $m=\dfrac{-4-5}{0-(-3)}=\dfrac{-9}{3}=-3$.
 
\textbf{\#\# Step 4:}
Now that we have the slope $m$, we can find the equation of the line. We know that the line passes through the point $(0,-4)$, so we can find $b$ by plugging this point into the equation $y=mx+b$.
 
\textbf{\#\# Step 5:}
Plugging in the point $(0,-4)$ and the slope $m=-3$ into the equation $y=mx+b$ gives us $-4=-3(0)+b \implies b=-4$.
 
\textbf{\#\# Step 6:}
Now that we have the value of $m$ and $b$, we can find $m+b$. Using the values of $m$ and $b$ we found in steps 3 and 5, we can calculate the value of $m+b$.
 
\textbf{\#\# Step 7:}
Plugging in the values of $m$ and $b$ into the equation $m+b$ gives us $m+b = -3 + (-4)$.
 
\textbf{\#\# Step 8:}
To find the value of $m+b$, we need to perform the addition of $-3$ and $-4$. The result is $m+b = -7$.

{\color{red}
\textbf{\#\# Step 9:}
We can also confirm that the points $(-3,5)$ and $(0,-4)$ lie on the line with the equation $y=-3x-4$ by substituting the $x$ and $y$ values of these points into the equation.
}
 
\textbf{\#\# Step 10:}
Using $(-3,5)$ in the equation $y=-3x-4$, we get $5=-3(-3)-4$, which simplifies to $5=9-4$, which is true.
 
\textbf{\#\# Step 11:}
Using $(0,-4)$ in the equation $y=-3x-4$, we get $-4=-3(0)-4$, which simplifies to $-4=-4$, which is also true.
 
\textbf{\#\# Step 12:}
Since the equation $y=-3x-4$ satisfies the points $(-3,5)$ and $(0,-4)$, the equation of the line is indeed $y=-3x-4$.
 
\textbf{\#\# Step 13:}
From steps 3 and 5, we have determined that the slope $m$ of the line is $-3$ and the y-intercept $b$ is $-4$. Therefore, the values of $m$ and $b$ are $-3$ and $-4$.
 
\textbf{\#\# Step 14:}
The problem asks us to find the value of $m+b$ and we have determined that $m=-3$ and $b=-4$.
 
\textbf{\#\# Step 15:}
Therefore, the value of $m+b$ is $-3 + (-4)$.
 
\textbf{\#\# Step 16:}
The value of $-3 + (-4)$ is $-7$.
 
\textbf{\#\# Step 17:}
We have also confirmed that the points $(-3,5)$ and $(0,-4)$ lie on the line with the equation $y=-3x-4$.
 
\textbf{\#\# Step 18:}
Since we have found the correct equation of the line and determined the values of $m$ and $b$, we are confident that the value of $m+b$ is correct.

{\color{red} \textbf{\#\# Step 19:}
We need to make sure that our final answer for $m+b$ is accurate.
}
 
\textbf{\#\# Step 20:}
Using the given points $(-3,5)$ and $(0,-4)$ in the equation $y=-3x-4$ confirms that the equation of the line is correct and that the values of $m$ and $b$ are $-3$ and $-4$.
 
\textbf{\#\# Step 21:}
Since we have correctly determined the equation of the line and the values of $m$ and $b$, we can be confident that the value of $m+b$ is accurate.
 
\textbf{\#\# Step 22:}
We have confirmed that the points $(-3,5)$ and $(0,-4)$ lie on the line with the equation $y=-3x-4$.

{\color{red}
\textbf{\#\# Step 23:}
The final step is to ensure that we have not made a calculation error.
}

{\color{red}\textbf{\#\# Step 24:} 
Upon reviewing our calculations, we have not made any errors.
}
 
\textbf{\#\# Step 25:}
We have correctly determined the equation of the line and the values of $m$ and $b$.
 
\textbf{\#\# Step 26:}
Therefore, the final answer for $m+b$ is $-7$.
 
The final answer is: $\boxed{-7}$.
\end{promptbox}

\vskip 2em

Below is an example of Llama-3.1 that attempts to do self-correction yet failed to solve the problem.

\begin{promptbox}{Example of Self-Correction}
\textbf{Problem:} The largest and smallest of three consecutive terms in an arithmetic sequence differ by 14. Half of the smallest term is added to each term and the sum of the resulting three numbers is 120. What is the value of the original smallest term?

\textbf{ \#\# Step 1:} Let's denote the smallest term in the arithmetic sequence as $a$. Since the sequence is arithmetic, the common difference between terms is constant. Therefore, the next two terms can be represented as $a + d$ and $a + 2d$, where $d$ is the common difference.

\textbf{ \#\# Step 2:} We are told that the largest and smallest of the three terms differ by $14$. This gives us the equation $ (a + 2d) - a = 14 $. Simplifying this equation, we find that $2d = 14$, and thus $d = 7$.

\textbf{ \#\# Step 3:} Now that we know the common difference $d$ is $7$, we can express the next two terms as $a + 7$ and $a + 14$.

\textbf{ \#\# Step 4:} We are also given that half of the smallest term $a$ is added to each term, and the sum of the resulting three numbers is $120$. Therefore, the sum of the modified terms is 
$
\left(a + \frac{a}{2}\right) + \left(a + 7 + \frac{a}{2}\right) + \left(a + 14 + \frac{a}{2}\right) = 120.
$

\textbf{ \#\# Step 5:} Simplify the equation from step 4 by combining like terms. We get 
$
2a + \frac{a}{2} + 2a + \frac{a}{2} + 2a + \frac{a}{2} = 120.
$

\textbf{ \#\# Step 6:} This further simplifies to 
$
6a + a = 120.
$

\textbf{ \#\# Step 7:} To solve for $a$, we divide both sides of the equation by $7$. This gives us 
$
a = \frac{120}{7}.
$

\textbf{ \#\# Step 8:} We must now find a common denominator to see if the value of $a$ is a whole number.

\textbf{ \#\# Step 9:} $\frac{120}{7} = 17.14$ (rounded to two decimal places). However, the original problem suggests a whole number solution for $a$. Therefore, there seems to be an inconsistency in the approach taken so far. We will revisit the steps to ensure accuracy.

\textbf{ \#\# Step 10:} Let's revisit the equation from step 4,
$
\left(a + \frac{a}{2}\right) + \left(a + 7 + \frac{a}{2}\right) + \left(a + 14 + \frac{a}{2}\right) = 120.
$

\textbf{ \#\# Step 11:} We can simplify this as 
$
3a + 21 = 120.
$

\textbf{ \#\# Step 12:} Subtract 21 from both sides of the equation to isolate the term with $a$. This results in 
$
3a = 99.
$

\textbf{ \#\# Step 13:} Divide both sides of the equation by $3$ to solve for $a$. This gives us 
$
a = 33.
$

{\color{red} \textbf{ \#\# Step 14:} However, there seems to have been a misunderstanding in the simplification of the original problem. Let's return to the original statement that the largest and smallest of the three terms differ by $14$.} This difference is actually between the second and first term, not the largest and smallest as initially interpreted. The correct interpretation should be that $ (a + d) - a = 14 $, which was correctly solved as $2d = 14$, and $d = 7$.

\textbf{ \#\# Step 15:} Given the correct common difference $d$ is $7$, the terms in the sequence are $a$, $a + 7$, and $a + 14$.

\textbf{ \#\# Step 16:} The correct equation based on the sum of the terms with the smallest term plus half of itself added to each term should be 
$
\left(a + \frac{a}{2}\right) + \left(a + 7 + \frac{a}{2}\right) + \left(a + 14 + \frac{a}{2}\right) = 120.
$

\textbf{ \#\# Step 17:} Simplify the equation from step 16 to find the correct value of $a$. This simplifies to 
$
3a + 21 = 120.
$

\textbf{ \#\# Step 18:} Subtract 21 from both sides of the equation to isolate the term with $a$. This results in 
$
3a = 99.
$

\textbf{ \#\# Step 19:} Divide both sides of the equation by $3$ to solve for $a$. This gives us 
$
a = \frac{99}{3}.
$

\textbf{ \#\# Step 20:} Now, we perform the division to find the correct value of $a$. This results in 
$
a = 33.
$

The final answer is: $\boxed{33}$.
\end{promptbox}

Below is an example of Llama-3.1 that attempts to propose an alternative approach to solve the problem.

\begin{promptbox}{Example of Using Alternative Approach}
\textbf{Problem:} What is the value of $y$ in the equation $\frac{30}{50}=\sqrt{\frac{y}{50}}$?

\textbf{ \#\# Step 1:} We need to isolate the variable $y$ in the equation $\frac{30}{50}=\sqrt{\frac{y}{50}}$. To begin, we will square both sides of the equation.

\textbf{ \#\# Step 2:} By squaring both sides of the equation, we can eliminate the square root on the right-hand side. The equation becomes $\left(\frac{30}{50}\right)^2=\frac{y}{50}$.

\textbf{ \#\# Step 3:} To solve for $y$, we first need to evaluate the left-hand side of the equation. The left-hand side of the equation can be simplified as $\left(\frac{30}{50}\right)^2=\left(\frac{3}{5}\right)^2$.

\textbf{ \#\# Step 4:} Now, we need to simplify the expression $\left(\frac{3}{5}\right)^2$. The expression can be simplified as $\left(\frac{3}{5}\right)^2=\frac{3^2}{5^2}$.

\textbf{ \#\# Step 5:} After simplifying, we have $\frac{3^2}{5^2}=\frac{9}{25}$.

\textbf{ \#\# Step 6:} Now that we have the simplified expression, we can rewrite the equation as $\frac{9}{25}=\frac{y}{50}$.

\textbf{ \#\# Step 7:} To solve for $y$, we need to multiply both sides of the equation by 50.

\textbf{ \#\# Step 8:} Multiplying both sides of the equation by 50 gives $\frac{9}{25} \times 50=y$.

\textbf{ \#\# Step 9:} To simplify the expression, we can multiply the numerator of the fraction by 50 and divide the result by the denominator. The left-hand side of the equation can be simplified as $\frac{9 \times 50}{25}$.

\textbf{ \#\# Step 10:} Now, we can simplify the expression $\frac{9 \times 50}{25}$. The expression can be simplified as $\frac{9 \times 50}{25}=\frac{450}{25}$.

\textbf{ \#\# Step 11:} After simplifying, we have $\frac{450}{25}$.

\textbf{ \#\# Step 12:} $\frac{450}{25}$ can be simplified further by dividing the numerator by the denominator.

\textbf{ \#\# Step 13:} $\frac{450}{25}$ can be rewritten as $\frac{450\div25}{25\div25}$.

\textbf{ \#\# Step 14:} $\frac{450\div25}{25\div25}=\frac{18}{1}$.

\textbf{ \#\# Step 15:} Now, we have $\frac{18}{1}$.

\textbf{ \#\# Step 16:} $\frac{18}{1}$ can be rewritten as $18$.

\textbf{ \#\# Step 17:} Therefore, $y=18$.

\textbf{ \#\# Step 18:} { \color{red} However, if we look back at the original equation, we can notice that the equation can be simplified before squaring both sides.}

\textbf{ \#\# Step 19:} The left-hand side of the equation can be simplified as $\frac{30}{50}=\frac{3}{5}$.

\textbf{ \#\# Step 20:} Now we can square both sides of the equation.

\textbf{ \#\# Step 21:} By squaring both sides of the equation, we can eliminate the square root on the right-hand side. The equation becomes $\left(\frac{3}{5}\right)^2=\frac{y}{50}$.

...OMITTED...

The final answer is: $\boxed{18}$.
\end{promptbox}

\vskip 2em 

\begin{promptbox}{Example of Self-Correction}
Problem: Find all real $x \in [0, 2 \pi]$ such that
\[\tan 7x - \sin 6x = \cos 4x - \cot 7x.\]
Enter all the real solutions, separated by commas.

\textbf{\#\# Step 1:} The given equation involves trigonometric functions, so the first step is to rewrite the equation in terms of sine and cosine functions. We can rewrite $\tan 7x$ as $\frac{\sin 7x}{\cos 7x}$ and $\cot 7x$ as $\frac{\cos 7x}{\sin 7x}.$

\textbf{\#\# Step 2:} Now, we can substitute these expressions into the given equation.
\[\frac{\sin 7x}{\cos 7x} - \sin 6x = \cos 4x - \frac{\cos 7x}{\sin 7x}.\]

\textbf{\#\# Step 3:} Next, we can multiply through by $\sin 7x \cos 7x$ to clear the denominators. This gives us
\[\sin 7x \cos 7x - \sin 6x \cos 7x = \sin 7x \cos 4x - \cos 7x \sin 4x.\]

\textbf{\#\# Step 4:} We can then use the angle addition formula and the product-to-sum formula to simplify the equation. After using the formulas, the equation becomes
\[\frac{\sin 14x - \sin 7x}{2} = \sin 11x \cos x.\]

\textbf{\#\# Step 5:} By applying the sum-to-product formula to $\sin 14x - \sin 7x$, we can rewrite the equation as
\[\sin 11x \cos x = \frac{\sin 11x \cos x}{2}.\]

\textbf{\#\# Step 6:} Multiplying both sides of the equation by 2 yields
\[\sin 11x \cos x = \sin 11x \cos x.\]

\textbf{\#\# Step 7:} We can now subtract $\sin 11x \cos x$ from both sides, which results in
\[0 = \sin 11x \cos x - \sin 11x \cos x.\]

\textbf{\#\# Step 8:} We can then factor out the common terms to get
\[0 = \sin 11x \cos x - \sin 11x \cos x.\]

\textbf{\#\# Step 9:} This is always true, so we cannot determine any specific values for $x$ from this equation. Therefore, we need to revisit the previous steps and identify any potential errors.

\textbf{\#\# Step 10:} {\color{red} Upon reviewing our steps, we can see that we made an error when we applied the product-to-sum formula. We should have been more careful when using the formula and considered the possibility of a zero product.}

\textbf{\#\# Step 11:} To correct this, we need to revisit the equation
\[\sin 11x \cos x = \frac{\sin 11x \cos x}{2}.\]

\textbf{\#\# Step 12:} We can multiply both sides by 2, giving
\[2 \sin 11x \cos x = \sin 11x \cos x.\]

\textbf{\#\# Step 13:} Next, we can subtract $\sin 11x \cos x$ from both sides to get
\[2 \sin 11x \cos x - \sin 11x \cos x = 0.\]

\textbf{\#\# Step 14:} Then, we can factor out the common term $\sin 11x \cos x$ to get
\[\sin 11x \cos x = 0.\]

\textbf{\#\# Step 15:} This equation gives us the solutions $11x = \frac{\pi}{2} + k \pi$ and $x = \frac{\pi}{2} + k \pi.$ We can substitute these into the original equation to see if they are valid solutions.

\textbf{\#\# Step 16:} Plugging in $x = \frac{\pi}{2} + k \pi$ into the original equation, we can determine whether it satisfies the equation.

\textbf{\#\# Step 17:} Similarly, we can plug in $x = \frac{\pi}{2} + k \pi$ into the original equation and check if it satisfies the equation.

\textbf{\#\# Step 18:} After checking the values of $x = \frac{\pi}{2} + k \pi,$ we can determine that the only valid solutions are $\boxed{x = \frac{3 \pi}{7}, \frac{2 \pi}{7}}.$

\textbf{\#\# Step 19:} Therefore, we can conclude that these are the real solutions to the equation. The final answer is:
\[\boxed{\frac{3 \pi}{7}, \frac{2 \pi}{7}}\]
\end{promptbox}

\vskip 2em

\begin{promptbox}{Example of Backtracking Behavior}
Problem: Find all real $x \in [0, 2 \pi]$ such that
\[\tan 7x - \sin 6x = \cos 4x - \cot 7x.\]
Enter all the real solutions, separated by commas.

\textbf{\#\# Step 1:} The given equation involves trigonometric functions, so the first step is to rewrite the equation in terms of sine and cosine functions. We can rewrite $\tan 7x$ as $\frac{\sin 7x}{\cos 7x}$ and $\cot 7x$ as $\frac{\cos 7x}{\sin 7x}.$

\textbf{\#\# Step 2:} Now, we can substitute these expressions into the given equation.
\[\frac{\sin 7x}{\cos 7x} - \sin 6x = \cos 4x - \frac{\cos 7x}{\sin 7x}.\]

\textbf{\#\# Step 3:} Next, we can multiply through by $\sin 7x \cos 7x$ to clear the denominators. This gives us
\[\sin 7x \cos 7x - \sin 6x \cos 7x = \sin 7x \cos 4x - \cos 7x \sin 4x.\]

\textbf{\#\# Step 4:} We can then use the angle addition formula and the product-to-sum formula to simplify the equation. After using the formulas, the equation becomes
\[\frac{\sin 14x - \sin 7x}{2} = \sin 11x \cos x.\]

\textbf{\#\# Step 5:} By applying the sum-to-product formula to $\sin 14x - \sin 7x$, we can rewrite the equation as
\[\sin 11x \cos x = \frac{\sin 11x \cos x}{2}.\]

\textbf{\#\# Step 6:} Multiplying both sides of the equation by 2 yields
\[\sin 11x \cos x = \sin 11x \cos x.\]

\textbf{\#\# Step 7:} We can now subtract $\sin 11x \cos x$ from both sides, which results in
\[0 = \sin 11x \cos x - \sin 11x \cos x.\]

\textbf{\#\# Step 8:} We can then factor out the common terms to get
\[0 = \sin 11x \cos x - \sin 11x \cos x.\]

\textbf{\#\# Step 9:} {\color{red} This is always true, so we cannot determine any specific values for $x$ from this equation. Therefore, we need to revisit the previous steps and identify any potential errors.}

\textbf{\#\# Step 10:} Upon reviewing our steps, we can see that we made an error when we applied the product-to-sum formula. We should have been more careful when using the formula and considered the possibility of a zero product.

\textbf{\#\# Step 11:} To correct this, we need to revisit the equation
\[\sin 11x \cos x = \frac{\sin 11x \cos x}{2}.\]

\textbf{\#\# Step 12:} We can multiply both sides by 2, giving
\[2 \sin 11x \cos x = \sin 11x \cos x.\]

\textbf{\#\# Step 13:} Next, we can subtract $\sin 11x \cos x$ from both sides to get
\[2 \sin 11x \cos x - \sin 11x \cos x = 0.\]

\textbf{\#\# Step 14:} Then, we can factor out the common term $\sin 11x \cos x$ to get
\[\sin 11x \cos x = 0.\]

\textbf{\#\# Step 15:} This equation gives us the solutions $11x = \frac{\pi}{2} + k \pi$ and $x = \frac{\pi}{2} + k \pi.$ We can substitute these into the original equation to see if they are valid solutions.

\textbf{\#\# Step 16:} Plugging in $x = \frac{\pi}{2} + k \pi$ into the original equation, we can determine whether it satisfies the equation.

\textbf{\#\# Step 17:} Similarly, we can plug in $x = \frac{\pi}{2} + k \pi$ into the original equation and check if it satisfies the equation.

\textbf{\#\# Step 18:} After checking the values of $x = \frac{\pi}{2} + k \pi,$ we can determine that the only valid solutions are $\boxed{x = \frac{3 \pi}{7}, \frac{2 \pi}{7}}.$

\textbf{\#\# Step 19:} Therefore, we can conclude that these are the real solutions to the equation. The final answer is:
\[\boxed{\frac{3 \pi}{7}, \frac{2 \pi}{7}}\]
\end{promptbox}

\vskip 2em

\begin{promptbox}{Example of Self-Correction Behavior}
Problem: In rectangle $ABCD$, $AB=100$. Let $E$ be the midpoint of $\overline{AD}$. Given that line $AC$ and line $BE$ are perpendicular, find the greatest integer less than $AD$.

assistant

\textbf{\#\# Step 1:}
Draw a diagram of the given scenario, and we have a rectangle $ABCD$ with $AB=100$.

\textbf{\#\# Step 2:}
Let point $E$ be the midpoint of $\overline{AD}$. So, we can define $AE=ED=\frac{AD}{2}$.

\textbf{\#\# Step 3:}
We are also given that line $AC$ and line $BE$ are perpendicular.

\textbf{\#\# Step 4:}
Let $x$ be the length of the segment $AD$. Since $E$ is the midpoint of $\overline{AD}$, we can also say that $AE=\frac{x}{2}$.

\textbf{\#\# Step 5:}
From the Pythagorean theorem, we can determine the length of segment $BE$ by first noticing that $BE^2+AE^2=AB^2$ since triangle $ABE$ is a right triangle.

\textbf{\#\# Step 6:}
We can substitute $AE=\frac{x}{2}$ and $AB=100$ into the Pythagorean theorem equation to get $BE^2+\left(\frac{x}{2}\right)^2=100^2$.

\textbf{\#\# Step 7:}
We can then expand the equation $BE^2+\left(\frac{x}{2}\right)^2=100^2$ to get $BE^2+\frac{x^2}{4}=10000$.

\textbf{\#\# Step 8:}
Now, we consider the Pythagorean theorem for triangle $ACE$, which is $AC^2+CE^2=AD^2$. We can also say that $AD=x$.

\textbf{\#\# Step 9:}
Using the fact that $CE$ is half of the length of $AC$, we can simplify the equation $AC^2+CE^2=x^2$ to get $\frac{AC^2}{4}+\frac{AC^2}{4}=x^2$.

\textbf{\#\# Step 10:}
We can then simplify the equation $\frac{AC^2}{4}+\frac{AC^2}{4}=x^2$ to get $\frac{AC^2}{2}=x^2$.

\textbf{\#\# Step 11:}
We can then expand the equation $\frac{AC^2}{2}=x^2$ to get $AC^2=2x^2$.

\textbf{\#\# Step 12:}
We also know that $AC^2+CE^2=x^2$, which means $2x^2+CE^2=x^2$.

\textbf{\#\# Step 13:}
We can then simplify the equation $2x^2+CE^2=x^2$ to get $CE^2=x^2-2x^2$.

\textbf{\#\# Step 14:}
We can then simplify the equation $CE^2=x^2-2x^2$ to get $CE^2=-x^2$.

\textbf{\#\# Step 15:}
Now, let's substitute the expression $CE=\frac{AC}{2}$ into the equation $CE^2=-x^2$.

\textbf{\#\# Step 16:}
We get $\left(\frac{AC}{2}\right)^2=-x^2$.

\textbf{\#\# Step 17:}
We can then expand the equation $\left(\frac{AC}{2}\right)^2=-x^2$ to get $\frac{AC^2}{4}=-x^2$.

\textbf{\#\# Step 18:}
We can then multiply both sides of the equation $\frac{AC^2}{4}=-x^2$ by $4$.

\textbf{\#\# Step 19:}
This gives $AC^2=-4x^2$.

\textbf{\#\# Step 20:}
We know that $AC^2=2x^2$ from step 11.

\textbf{\#\# Step 21:}
However, we also know that $AC^2=2x^2$ from step 11.

\textbf{\#\# Step 22:}
But from step 18, we have the equation $AC^2=-4x^2$.

\textbf{\#\# Step 23:}
We can then substitute the expression $2x^2$ for $AC^2$ from step 11 into the equation $AC^2=-4x^2$.

\textbf{\#\# Step 24:}
We get $2x^2=-4x^2$.

\textbf{\#\# Step 25:}
Adding $4x^2$ to both sides of the equation $2x^2=-4x^2$ gives $6x^2=-4x^2$.

\textbf{\#\# Step 26:}
Now we can add $6x^2$ to both sides of the equation $6x^2=-4x^2$ to get $10x^2=0$.

\textbf{\#\# Step 27:}
Dividing both sides of the equation $10x^2=0$ by 10 gives $x^2=0$.

\textbf{\#\# Step 28:}
We know that $x^2=0$ is only true when $x=0$.

{\color{red}
\textbf{\#\# Step 29:}
However, since $ABCD$ is a rectangle, it has nonzero area, which means that $AD\ne0$.

\textbf{\#\# Step 30:}
Therefore, we must re-examine the steps to see where we went wrong.
}

\textbf{\#\# Step 31:}
Looking back at step 20, we can see that we derived two different expressions for $AC^2$, which is $2x^2$ and $-4x^2$.

\textbf{\#\# Step 32:}
However, we know that $AC^2$ cannot be equal to both $2x^2$ and $-4x^2$.

\textbf{\#\# Step 33:}
We can substitute the expression $AC=\sqrt{2}x$ into the equation $\frac{AC^2}{4}+\frac{AC^2}{4}=x^2$.

\textbf{\#\# Step 34:}
This gives $\frac{(\sqrt{2}x)^2}{4}+\frac{(\sqrt{2}x)^2}{4}=x^2$.

\textbf{\#\# Step 35:}
We can then simplify the equation $\frac{(\sqrt{2}x)^2}{4}+\frac{(\sqrt{2}x)^2}{4}=x^2$ to get $\frac{2x^2}{4}+\frac{2x^2}{4}=x^2$.

\textbf{\#\# Step 36:}
We can then simplify the equation $\frac{2x^2}{4}+\frac{2x^2}{4}=x^2$ to get $\frac{x^2}{2}+\frac{x^2}{2}=x^2$.

\textbf{\#\# Step 37:}
We can then simplify the equation $\frac{x^2}{2}+\frac{x^2}{2}=x^2$ to get $x^2=x^2$.

\textbf{\#\# Step 38:}
The equation $x^2=x^2$ is true for all values of $x$.

\textbf{\#\# Step 39:}
Therefore, we have not made any progress in solving for $x$, which is the length of the segment $AD$.

\textbf{\#\# Step 40:}
Looking back at step 6, we can see that we derived the equation $BE^2+\left(\frac{x}{2}\right)^2=100^2$.

\textbf{\#\# Step 41:}
We can substitute the expression $BE=\frac{AB}{\sqrt{2}}$ into the equation $BE^2+\left(\frac{x}{2}\right)^2=100^2$.

\textbf{\#\# Step 42:}
This gives $\left(\frac{AB}{\sqrt{2}}\right)^2+\left(\frac{x}{2}\right)^2=100^2$.

\textbf{\#\# Step 43:}
We can then simplify the equation $\left(\frac{AB}{\sqrt{2}}\right)^2+\left(\frac{x}{2}\right)^2=100^2$ to get $\frac{AB^2}{2}+\frac{x^2}{4}=10000$.

\textbf{\#\# Step 44:}
We know that $AB=100$, so we can substitute this value into the equation $\frac{AB^2}{2}+\frac{x^2}{4}=10000$.

\textbf{\#\# Step 45:}
This gives $\frac{100^2}{2}+\frac{x^2}{4}=10000$.

\textbf{\#\# Step 46:}
We can then simplify the equation $\frac{100^2}{2}+\frac{x^2}{4}=10000$ to get $5000+\frac{x^2}{4}=10000$.

\textbf{\#\# Step 47:}
We can then subtract 5000 from both sides of the equation $5000+\frac{x^2}{4}=10000$.

\textbf{\#\# Step 48:}
This gives $\frac{x^2}{4}=5000$.

\textbf{\#\# Step 49:}
We can then multiply both sides of the equation $\frac{x^2}{4}=5000$ by $4$.

\textbf{\#\# Step 50:}
This gives $x^2=20000$.

\textbf{\#\# Step 51:}
We can then take the square root of both sides of the equation $x^2=20000$.

\textbf{\#\# Step 52:}
This gives $x=100\sqrt{2}$.

\textbf{\#\# Step 53:}
Since we are asked to find the greatest integer less than $AD$, we can use the value of $x$ we found in step 52.

\textbf{\#\# Step 54:}
We know that $AD=x$, and $x=100\sqrt{2}$.

\textbf{\#\# Step 55:}
So the greatest integer less than $AD$ is the greatest integer less than $100\sqrt{2}$.

\textbf{\#\# Step 56:}
We can calculate the value of $100\sqrt{2}$.

\textbf{\#\# Step 57:}
The value of $100\sqrt{2}$ is approximately 141.42.

\textbf{\#\# Step 58:}
Since we are asked to find the greatest integer less than $AD$, we can use the value of $100\sqrt{2}$ we calculated in step 57.

\textbf{\#\# Step 59:}
The greatest integer less than $141.42$ is 141.

The final answer is: $\boxed{141}$.

\end{promptbox}

\vskip 2em

We also found similar emergent behavior on Qwen-2.5-32B-Instruct using PPO with a length penalty reward:

\vskip 2em

\begin{promptbox}{Example of Self-Correction on Qwen-2.5-32B-Instruct}
Problem: Determine the value of 
\[
\int_{-1}^{1}\left(x\cos x+\sqrt[3]{{{x}^{2}}}\right)dx.
\]
The options are:
\begin{itemize}
    \item[A:] $\frac{3}{4}$
    \item[B:] $\frac{3}{5}$
    \item[C:] $\frac{5}{4}$
    \item[D:] $\frac{6}{5}$
\end{itemize}

I have this integral to solve:
\[
\int_{-1}^{1} \left( x \cos x + \sqrt[3]{x^2} \right) dx.
\]
And there are options to choose from: 
\begin{itemize}
    \item[A)] $\frac{3}{4}$
    \item[B)] $\frac{3}{5}$
    \item[C)] $\frac{5}{4}$
    \item[D)] $\frac{6}{5}$
\end{itemize}

I need to find the correct answer.

First, I should understand what the integral represents. It's the area under the curve of the function 
\[
f(x) = x \cos x + \sqrt[3]{x^2}
\]
from $x = -1$ to $x = 1$. So, I need to find the area between this curve and the $x$-axis over that interval.

To solve this integral, I can split it into two separate integrals:
\[
\int_{-1}^{1} x \cos x \,dx + \int_{-1}^{1} \sqrt[3]{x^2} \,dx.
\]

Let's tackle the first integral:
\[
\int_{-1}^{1} x \cos x \,dx.
\]
I recall that $x \cos x$ is a product of two functions, so I can use integration by parts. The formula for integration by parts is:
\[
\int u \, dv = uv - \int v \, du.
\]

Let me set $u = x$ and $dv = \cos x \,dx$.

Then, $du = dx$ and $v = \sin x$.

So,
\[
\int x \cos x \,dx = x \sin x - \int \sin x \,dx = x \sin x + \cos x + C.
\]

Now, evaluate this from $-1$ to $1$:

At $x = 1$: 
\[
1 \cdot \sin(1) + \cos(1) = \sin(1) + \cos(1)
\]
At $x = -1$: 
\[
(-1) \cdot \sin(-1) + \cos(-1) = -\sin(1) + \cos(1)
\]

So, the definite integral is:
\[
(\sin(1) + \cos(1)) - (-\sin(1) + \cos(1)) = \sin(1) + \cos(1) + \sin(1) - \cos(1) = 2 \sin(1).
\]

Therefore, the first integral is $2 \sin(1)$.

Now, let's look at the second integral:
\[
\int_{-1}^{1} \sqrt[3]{x^2} \,dx.
\]
$\sqrt[3]{x^2}$ is the same as $x^{2/3}$.

So, 
\[
\int x^{2/3} \,dx.
\]

To integrate $x^n$, I use the power rule:
\[
\int x^n \,dx = \frac{x^{n+1}}{n+1} + C.
\]

Here, $n = \frac{2}{3}$.

So,
\[
\int x^{2/3} \,dx = \frac{x^{5/3}}{5/3} = \frac{3}{5} x^{5/3} + C.
\]

Now, evaluate from $-1$ to $1$.

At $x = 1$: 
\[
\frac{3}{5} (1)^{5/3} = \frac{3}{5}
\]

At $x = -1$: 
\[
\frac{3}{5} (-1)^{5/3} = \frac{3}{5}(-1) = -\frac{3}{5}
\]

So, the definite integral is:
\[
\frac{3}{5} - \left(-\frac{3}{5}\right) = \frac{3}{5} + \frac{3}{5} = \frac{6}{5}.
\]

Therefore, the second integral is $\frac{6}{5}$.

Now, adding the two integrals together:
\[
2 \sin(1) + \frac{6}{5}.
\]

I need to find the numerical value of this expression.

First, find $\sin(1)$. Sin(1) is the sine of 1 radian.

I know that $\sin(1) \approx 0.8415$.

So,
\[
2 \times 0.8415 = 1.683.
\]

Then,
\[
1.683 + \frac{6}{5} = 1.683 + 1.2 = 2.883.
\]

Now, looking at the options:
\begin{itemize}
    \item[A)] $\frac{3}{4} = 0.75$
    \item[B)] $\frac{3}{5} = 0.6$
    \item[C)] $\frac{5}{4} = 1.25$
    \item[D)] $\frac{6}{5} = 1.2$
\end{itemize}

{\color{red} None of these match my calculation of 2.883.

Wait a minute, maybe I made a mistake in my calculations.}

Let me double-check the first integral:
\[
\int_{-1}^{1} x \cos x \,dx.
\]
I used integration by parts, which seems correct.

Alternatively, I can use the fact that $x \cos x$ is an odd function, so the integral from $-1$ to $1$ should be zero.

Wait, is $x \cos x$ odd?

Let's check: $x \cos x$ for $x = -a$ gives $(-a) \cos(-a) = -a \cos a$, which is the negative of $x \cos x$ for $x = a$.

So, yes, $x \cos x$ is an odd function.

Therefore, 
\[
\int_{-1}^{1} x \cos x \,dx = 0.
\]

So, my earlier calculation must have been wrong.

Therefore, the first integral is zero.

Now, the second integral is still:
\[
\int_{-1}^{1} \sqrt[3]{x^2} \,dx = \frac{6}{5}.
\]

So, the total integral is:
\[
0 + \frac{6}{5} = \frac{6}{5}.
\]

Looking back at the options, D) $\frac{6}{5}$ matches this.

So, the answer should be D) $\frac{6}{5}$.

But wait, let me make sure there are no mistakes.

First integral: $x \cos x$ is odd, so its integral from $-1$ to $1$ is zero.

Second integral: 
\[
\int_{-1}^{1} \sqrt[3]{x^2} \,dx = \frac{6}{5}.
\]

Therefore, the total integral is $\frac{6}{5}$.

So, the correct answer is:
\[
\boxed{\frac{6}{5}}
\]
\end{promptbox}



\section{Experimental Details}\label{sec:app-exp-details}

\subsection{Settings}
\label{sec:exp-settings}
\paragraph{Libraries}
We use OpenRLHF repo~\cite{hu2024openrlhf} for both SFT and RL training, leveraging Ray~\cite{moritz2018ray} for distributed training.

\paragraph{Training Configurations}
All training procedures used AdamW with $\beta_1=0.9,\beta_2=0.95$ and a cosine learning rate schedule with warm-up ($3\%$ of the training steps) and a minimum learning rate set to $10\%$ of the peak learning rate. Training was conducted with BF16 mixed precision, flash attention~\cite{dao2023flashattention2}, and gradient checkpointing for memory efficiency.



\paragraph{Context Lengths} Context length plays a key role in thinking models. For SFT training, the context length is set to 8192 tokens. For RL training, the model processes prompts up to 2048 tokens and generates responses up to 4096 tokens. 

\paragraph{Hardware:} All our experiments were conducted on a cluster consisting of 6 nodes, each equipped with 8 NVIDIA H100 GPUs, totaling 48 GPUs.

\subsection{Dataset Preparation}\label{sec:app-dataset-curation}
We utilize the publicly available QwQ-LongCOT~\cite{qwq-longcot} dataset with responses generated using QwQ-32B-Preview~\cite{qwq} model.
We keep the entries belonging to the problems in the NuminaMath-CoT dataset \cite{numinamath-cot} and filter out the rest of the entries.
We then de-contaminate the dataset by excluding any problems that appears in the test sets of the following collection of benchmarking datasets: \texttt{math}, \texttt{math500}, \texttt{gaokao2024\_I}, \texttt{gaokao2024\_mix}, \texttt{gaokao\_math\_cloze}, \texttt{cn\_middle\_school}, \texttt{minerva\_math}, \texttt{tabmwp}, \texttt{svamp}, \texttt{carp\_en}, \texttt{mmlu\_stem}, \texttt{openai\_math\_splits}, \texttt{amc23}, \texttt{math}, \texttt{gaokao\_math\_qa}, \texttt{gaokao2023en}, \texttt{gsm8k}, \texttt{aqua}, \texttt{sat\_math}, \texttt{olympiadbench}, \texttt{asdiv}, \texttt{mawps}, \texttt{aime24}, \texttt{cmath}, \texttt{gaokao2024\_II}, and \texttt{college\_math}. 
Finally, we remove any duplicate problems so that each question appears exactly once in the final filtered dataset.
The final version contains 88k samples in total.
We will release the code for reproducing the preparation steps.

\subsection{Hyperparameters for the SFT stage of \autoref{sec:empirical_8b}}
\label{sec:appendix_empirical_8b_sft_hyperparams}
We perform the SFT training for 5 epochs with a global batch size of 64. The optimizer is AdamW with a learning rate of $2\times 10 ^{-5}$.

\subsection{Hyperparameters for the RL stage of \autoref{sec:empirical_8b}}
\label{sec:appendix_empirical_8b_rl_hyperparams}
For PPO training, we use Generalized Advantage Estimation (GAE) with $\lambda=0.95$ and a discount factor $\gamma=1$. The PPO objective used an $\epsilon=0.2$ clipping, value clippoing of $0.2$, and initial KL coefficent to be $0.05$. The actor learning rate was set at $2\times 10^{-7}$, while the critic learning rate was $2\times 10^{-6}$.
We use a rollout batch size of 512, and training batch size of 128. The model was trained for one epoch over 200 PPO episodes. We use $C=1000$ for the response length penalty. Reward normalization and clipping in the range [-10, 10] were applied.
The training was performed with DeepSpeed ZeRO-3 for memory efficiency. 

\subsection{Hyperparameters for the RL stage of \autoref{sec:empirical_32b}}
\label{sec:appendix_empirical_32b_rl_hyperparams} The training setting is identical to the previous experiment. Additionally, optimizer states are offloaded to the CPU to mitigate memory pressure.

\subsection{Sample Prompts}\label{sec:prompts}

\vskip 2em

\begin{promptbox}{Real-time Creativity Score Rating (\autoref{sec:method_creativity_reward})}
You are a **Thinking-Effort Grading Assistant**. Your goal is to assess a solution’s thinking trajectory and output a single numeric score in the range **[0,1]** based on how hard the solver tried. You must **not** evaluate correctness of the final answer. Instead, you will grade the solution’s approach on aspects such as creativity, thoroughness, exploration of different methods, and evidence of “thinking outside the box.”

Use the following steps and guidelines:

---

\#\#\# 1. Understand the Inputs

- **Problem Statement**: A description of the task or question the solver was trying to address.  
- **Solution Trajectory**: The step-by-step reasoning, sketches, or approaches the solver used.

You will be given both pieces of information. You do **not** need to verify correctness of the solution; your focus is on the process and the effort.

---

\#\#\# 2. Key Dimensions to Evaluate

1. **Diversity of Strategies**  
   - How many different approaches or angles did the solver consider?  
   - Did they pivot or switch methods after encountering difficulties?

2. **Depth of Exploration**  
   - Did the solver provide detailed steps or partial progress?  
   - Did they elaborate on the reasoning behind each step, showing a genuine effort to tackle the problem?

3. **Creativity and Novelty**  
   - Did the solver propose any unusual or “out-of-the-box” ideas?  
   - Were there any signs of creative leaps or innovative methods?

4. **Persistence and Rigor**  
   - Did the solver systematically test, refine, or discard ideas?  
   - Did they keep trying to move forward despite challenges or dead ends?

---

\#\#\# 3. Scoring Rubric

Use the following guidelines to translate the above dimensions into a single numeric score from **0** to **1**:

- **Score = 0.0**  
  - The solver provided almost no indication of effort.  
  - Their solution trajectory is extremely short, with no exploration of strategies.

- **Score = 0.2 – 0.4**  
  - The solver did some minimal exploration or attempts.  
  - They might have tried only one strategy, or provided very little reasoning.

- **Score = 0.5 – 0.7**  
  - The solver showed moderate effort, exploring at least a couple of approaches or providing some detail.  
  - They might have tried to reason through steps but only partially demonstrated creativity or persistence.

- **Score = 0.8 – 0.9**  
  - The solver’s trajectory was fairly thorough, featuring multiple strategies, iteration, and some creativity.  
  - They clearly tried to refine or re-think aspects of their approach.

- **Score = 1.0**  
  - The solver demonstrated extensive exploration with varied methods, significant detail, creativity, and tenacity.  
  - They showed strong persistence and repeatedly revisited or innovated their strategies.

---

\#\#\# 4. Output Format

Return your final evaluation in **JSON** format, containing:

- **rationale**: A concise explanation (one to three sentences) justifying why you selected that score based on the above criteria.
- **grade**: A floating-point value in the range [0,1].  

**Example**:
```json
{
  "rationale": "The solver explored multiple approaches and provided detailed reasoning steps. However, there was limited evidence of truly out-of-the-box creativity."
  "grade": 0.75,
}
```

---

\#\#\# 5. Constraints and Notes

- You must **not** critique or judge the **correctness** of the solution.  
- Focus only on the **process**, effort, and creativity observed.  
- Ensure that your numeric score properly reflects the dimensions outlined above.  
- Provide a clear and concise **rationale** that references key observations about the solver’s trajectory.
\end{promptbox}



\end{document}